\documentclass{article}

\usepackage[final,nonatbib]{neurips_data_2023}

\usepackage[numbers]{natbib}
\usepackage{bibunits}
\defaultbibliographystyle{abbrv}
\usepackage[utf8]{inputenc} %
\usepackage[T1]{fontenc}    %
\usepackage{url}            %
\usepackage{booktabs}       %
\usepackage{amsfonts}       %
\usepackage{nicefrac}       %
\usepackage{microtype}      %
\usepackage{graphicx}
\usepackage[dvipsnames]{xcolor}         %
\usepackage{changepage}
\usepackage{subcaption}
\usepackage{multirow}
\usepackage{comment}
\usepackage{rotating}
\usepackage{framed}
\usepackage{soul}
\usepackage{booktabs}
\usepackage{fontawesome}
\usepackage{fvextra}
\usepackage{tabularx}

\usepackage{hyperref}       %
\usepackage{cleveref}

\definecolor{formalshade}{rgb}{0.8, 0.9, 0.8}
\newenvironment{formal}{%
  \MakeFramed{\advance\hsize-\width\FrameRestore}%
  \noindent\hspace{-4.55pt}%
  \begin{adjustwidth}{}{7pt}%
  \vspace{0.5pt}\vspace{0.5pt}%
}
{%
  \vspace{0.5pt}\end{adjustwidth}\endMakeFramed%
}

\definecolor{mydarkblue}{rgb}{0,0.08,0.45}
\definecolor{mydarkgreen}{rgb}{0,0.45,0.15}
\hypersetup{colorlinks,linkcolor=red,urlcolor=mydarkblue,linktoc=page,citecolor=mydarkgreen}

\usepackage{pifont}%
\newcommand{\cmark}{\ding{51}}%
\newcommand{\xmark}{\ding{55}}%

\title{BEDD: The MineRL BASALT \\ Evaluation and Demonstrations Dataset \\ for Training and Benchmarking Agents \\ that Solve Fuzzy Tasks}

\author{%
  Stephanie Milani \\
  Carnegie Mellon University\\
  \texttt{smilani@cs.cmu.edu}
  \and
  Anssi Kanervisto\\
  Microsoft Research\\ 
  \texttt{anssi.kanervisto@microsoft.com}\\
  \and 
  Karolis Ramanauskas\\
  University of Bath\\
  \texttt{kr711@bath.ac.uk}\\
  \and
  Sander Schulhoff\\
  University of Maryland\\
  \texttt{sschulho@umd.edu}\\ 
  \and 
  Brandon Houghton\\
  OpenAI\\
  \texttt{brandon@openai.com}\\ 
  \and
  Rohin Shah\\
  \texttt{rohinmshah@gmail.com}\\ 
}

\usepackage{tikz}
\usepackage{footmisc}
\begin{document}

\maketitle

\newcommand{\change}[1]{\textcolor{black}{#1}}

\newcommand{\smnote}[1]{#1}
\newcommand{\smchange}[1]{#1}
\newcommand{\bhnote}[1]{\textcolor{Purple}{#1}}
\newcommand{\rs}[1]{\textcolor{blue}{RS: #1}}
\newcommand{\krnote}[1]{\textcolor{BurntOrange}{#1}}

\newcommand{\ak}[1]{\textcolor{purple}{AK: #1}}
\newcommand{\aknote}[1]{\textcolor{purple}{#1}}

\newcommand{\cavetasknospace}{\texttt{FindCave}}
\newcommand{\cavetask}{\cavetasknospace\ }
\newcommand{\waterfalltasknospace}{\texttt{MakeWaterfall}}
\newcommand{\waterfalltask}{\waterfalltasknospace\ }
\newcommand{\housetaskfull}{\texttt{BuildVillageHouse}\ }
\newcommand{\housetasknospace}{\texttt{BuildVillageHouse}}
\newcommand{\housetask}{\housetasknospace\ }
\newcommand{\pentaskfull}{\texttt{CreateVillageAnimalPen}\ }
\newcommand{\pentaskfullnospace}{\texttt{CreateVillageAnimalPen}}
\newcommand{\pentasknospace}{\texttt{AnimalPen}}
\newcommand{\pentask}{\texttt{AnimalPen}\ }
\newcommand{\shoveltask}{\texttt{ObtainDiamondShovel}\ }
\newcommand{\diamondtask}{\texttt{ObtainDiamond}\ }

\newcommand{\fulldataset}{\texttt{BEDD}}
\newcommand{\demodataset}{\texttt{Demonstrations Dataset}}
\newcommand{\evaldataset}{\texttt{Evaluation Dataset}}

\newcommand{\humanevalmujocometric}{TrueSkill?}

\newcommand{\demodatasetnot}{\mathcal{D}_{D}}
\newcommand{\evaldatasetnot}{\mathcal{D}_{E}}

\newcommand{\todo}[1]{\textcolor{red}{TODO: #1}}
\begin{abstract}

The MineRL BASALT competition has served to catalyze advances in learning from human feedback through four hard-to-specify tasks in Minecraft, such as \textit{create and photograph a waterfall}.
Given the completion of two years of BASALT competitions, we offer to the community a formalized benchmark through the BASALT Evaluation and Demonstrations Dataset (BEDD), which serves as a resource for algorithm development and performance assessment.
BEDD consists of a collection of 26 million image-action pairs from nearly 14,000 videos of human players completing the BASALT tasks in Minecraft.
It also includes over 3,000 dense pairwise human evaluations of human and algorithmic agents.
These comparisons serve as a fixed, preliminary leaderboard for evaluating newly-developed algorithms. %
To enable this comparison, we present a streamlined codebase for benchmarking new algorithms against the leaderboard.
In addition to presenting these datasets, we conduct a detailed analysis of the data from both datasets to guide algorithm development and evaluation. 
\change{The released code and data are available at \url{https://github.com/minerllabs/basalt-benchmark}.}

\end{abstract}

\section{Introduction}
\label{sec:intro}

\smchange{In traditional reinforcement learning, an agent learns how to act using reward based on an explicitly-defined reward signal~\citep{sutton2018reinforcement}.
This reward signal is often carefully designed by domain experts to communicate the intended goal for the agent to accomplish. 
Precisely specifying this form of feedback programmatically requires designers to \textit{a priori} enumerate all potential outcomes or constraints on how they would like the task to be completed. 
This enumeration is difficult to achieve in practice, and the resulting reward signals often fall short at correctly specifying the designer's intent \citep{krakovna2020specification}.
To address this challenge, researchers have explored the idea of incorporating alternative channels for communicating information about the desired behavior of the agent. 
This class of techniques is generally called \textit{learning from human feedback} (LfHF) \citep{christiano2017deep,jeon2020reward}.
The goal of LfHF is to utilize the feedback modalities most likely to result in an agent acting according to human-desired specifications.
}

\begin{figure}[t]
    \centering
    \includegraphics[width=.78\textwidth]{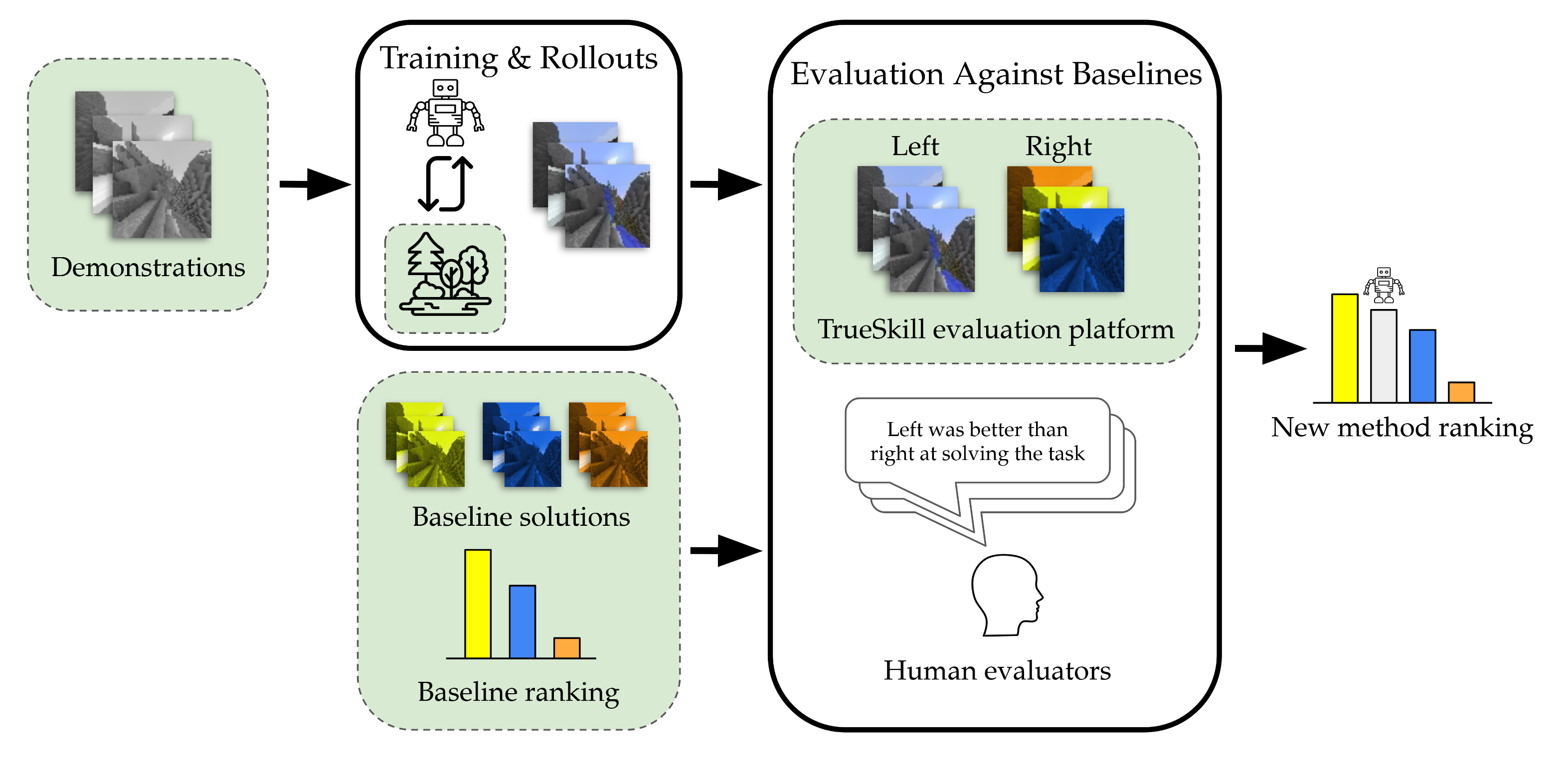}
    \caption{The BASALT benchmark. Green parts show the components of the included datasets. The agents are compared using video recordings of them solving the task. We contribute data for training agents to perform fuzzy tasks and for evaluating agents with human evaluators. We contribute code for further benchmarking.}
    \label{fig:overview}
\end{figure}

\smchange{
The complexity of this approach has been exemplified in the MineRL BASALT competitions~\citep{shah2021minerl}, which leverage the open-ended and flexible Minecraft environment to promote the development of LfHF algorithms. %
This competition series has provided a platform for developing agents capable of solving fuzzy tasks that lack well-defined reward signals. 
However, despite two years of competition, no agents have yet matched human performance levels~\citep{milani2023towards}, emphasizing the need to establish the BASALT tasks as a consistent, standardized benchmark.
In this paper, we clarify the BASALT benchmark and present concrete evaluation recommendations toward the goal of consistency.
}

\smchange{
To support this benchmark we introduce the BASALT Evaluation and Demonstrations Dataset (BEDD), an open and accessible dataset for learning to solve fuzzy tasks from human feedback. 
As shown in~\Cref{fig:overview}, this dataset consists of three main ingredients: the \demodataset, the \evaldataset, and supporting code for utilizing and analyzing the data. 
The \demodataset{} consists of over 26 million image-action pairs from 14,000 videos of labeled Minecraft gameplay of human players completing the BASALT tasks.
To facilitate evaluating agents using real human judgments, we present the \evaldataset, derived from the most recent BASALT competition~\citep{kanervisto2022basalt}. 
The \evaldataset{} consists of over 3,000 dense pairwise human evaluations of videos of various agents performing the BASALT tasks.
By dense, we mean that each evaluation includes a comparison of the relative task-completion performance of the agents and at least four additional questions, such as which agent was more human-like.
This results in 27,905 comparison points between $17$ different agents.
This dataset also includes a natural language response justifying why an agent was selected as being the better of the two.
The provided data functions as a leaderboard, offering researchers the ability to compare their newly-developed algorithm against various agents without redoing all costly human evaluations from scratch.} %

To facilitate the use of the \demodataset{} and the \evaldataset, we present a streamlined codebase. 
With this codebase, one can train a new model from the demonstration dataset and evaluate it against the provided leaderboard.
Alongside the presentation of these datasets, we conduct a detailed analysis of the data to guide algorithm development and evaluation. 
We release the code and detailed documentation for others to perform such analyses.
We hope that this codebase will assist others with quickly developing and evaluating LfHF algorithms to spur further progress toward agents that are better aligned with human intent. %

\section{The MineRL BASALT Benchmark}
\label{sec:basalt}
\begin{figure}[t]
  \centering
  \begin{subfigure}[b]{0.47\textwidth}
    \includegraphics[width=\textwidth]{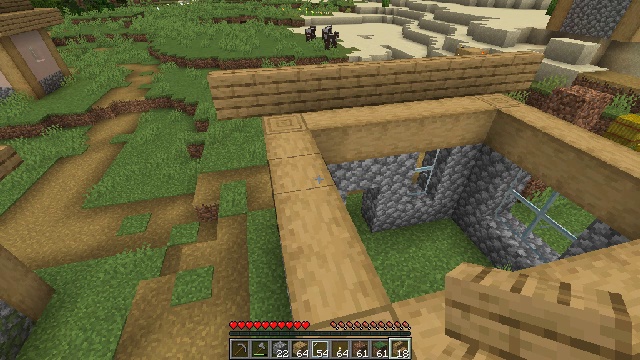}
  \end{subfigure}
  \begin{subfigure}[b]{0.47\textwidth}
    \includegraphics[width=\textwidth]{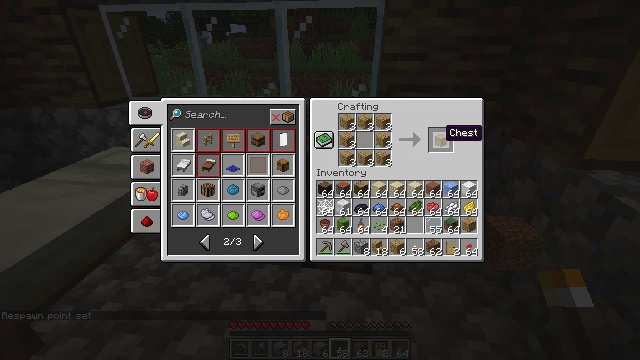}
  \end{subfigure} \\
  \caption{Two example images of the pixel observations provided by the MineRL environment used by BASALT. The agent receives pixels as observations and must use regular keyboard and mouse actions, including navigating the crafting menu with a mouse to make items.}
  \label{fig:MC-screenshots}
\end{figure}

We provide an overview of the MineRL BASALT benchmark, consisting of a task suite and evaluation framework for learning from human feedback, illustrated in (\Cref{fig:overview}). The benchmark uses Minecraft, a videogame that provides a rich and complex environment in which to define different tasks. The states are pixel observations; the actions are regular keyboard and mouse actions, closely following how humans play the game. This includes navigating the crafting menus using a mouse (\Cref{fig:MC-screenshots}).

\subsection{Tasks}
\label{subsec:tasks}
\smchange{All tasks are accompanied by a Gym environment~\citep{brockman2016openai} and either a simple English-language task description or a reward function to indicate the desired agent behavior.
The \textit{reward-free} tasks have the language task specification; the \textit{rewarded} tasks are accompanied by a reward function.}

\smchange{We use the same four reward-free tasks from the BASALT competitions, since these tasks were empirically validated to be challenging~\citep{milani2023towards,shah2022retrospective}.
}
\smchange{The goal of \cavetask is for the agent to discover a naturally-generated cave.}
\smchange{To complete \waterfalltasknospace, the agent must create a waterfall and subsequently take a picture of it. Because taking a photograph is not supported in Minecraft, we simulate it by providing the agent with an option to end the episode: the state at the moment that the agent ends the episode is its photograph.}
\smchange{In \pentaskfullnospace, the agent must build an animal pen next to an existing house, then corral a matching pair of farm animals into the pen. 
}
\smchange{In \housetasknospace, the agent spawns in a village and must build a new house in the same style as the other surrounding houses -- without damaging the village.
For more task details, see \Cref{sec:benchmark_appx}.}

\smchange{Because reward-free tasks are challenging to evaluate, we include \change{\texttt{ObtainDiamondShovel}~\citep{milani2023towards}, a task with a concrete reward function,} to enable quick iteration.
\change{This task} is more challenging than \texttt{ObtainDiamond}~\citep{guss2019neurips}, a well-established benchmark in reinforcement learning, since 
it requires an additional crafting step after obtaining the diamond.} An agent receives reward %
each time it obtains the next required resource in the crafting tree towards a diamond shovel.

\subsection{Evaluation}
\label{subsec:evaluation}
To fairly compare algorithms on a benchmark, \change{the parameters of that benchmark must be concretely defined. }%
\change{Previously}, many techniques claimed to solve MineRL \texttt{ObtainDiamond}. Often, \change{these successes involved} giving the agent access to the full game state instead of pixel observations, building in human priors \change{through} extensive action shaping, or \change{simplifying} the underlying environment dynamics. \change{Changes like these should be} clearly stated and applied equally to compared methods.
To avoid such \change{discrepancies} in the future, we specify some methodological best practices when using the BASALT benchmark: i) only pixel observations provided by the four environments should be used, ii) the environments should not be modified in any way, iii) action shaping~\cite{kanervisto2020action} is \change{permitted}, as long as it \change{applied to all methods}, iv) algorithms should be evaluated with TrueSkill~\citep{herbrich2006trueskill} using the specific hold-out test seeds provided in \Cref{sec:benchmark_appx}, and v) \smchange{the final evaluations must be conducted with human evaluators.}
For \change{further} guidance on how to fairly compare algorithms, we refer the reader to \change{recent work} \cite{patterson2023empirical}.

\smchange{BASALT agents are evaluated on hold-out test environments to produce multiple videos of agents attempting to complete that task. 
A single trial consists of two agents pitted against one another, which is shown to a human judge. 
This judge determines which agent completed the task better.}
\smchange{
The resulting dataset of human evaluations is assessed using the TrueSkill system, which%
} \smchange{
dynamically assesses the skills of a particular player (in this case, an LfHF algorithm). 
In addition to estimating the relative skill of an algorithm, it provides an uncertainty estimation. 
For a more extensive discussion of this procedure and a comparison of TrueSkill to other rating systems, please see~\Cref{sec:benchmark_appx}.
We hope that our demonstrations of successful task completions and the benchmarked algorithms from the competition can serve as a starting point for evaluating the performance of other algorithms.}

\smchange{
Developing algorithms while leveraging real human feedback is both expensive for machine learning practitioners and time-consuming for human evaluators.
Automated human evaluations have emerged as a valuable part of the pipeline for assessing various aspects of machine learning models~\citep{chen2022use,devlin2021navigation}. 
The goal is not to replace human evaluations; instead, it complements the process by providing quick, iterative feedback for algorithm development and initial assessment. 
As a result, we propose \textit{automating evaluations}~\citep{fan2022minedojo} as an additional component of the BASALT benchmark. 
The inclusion of \texttt{ObtainDiamondShovel} may help develop reward modeling techniques~\citep{ibarz2018reward} due to its associated concrete reward function. 
More generally, we hope that with the release of the \evaldataset, others can begin developing approaches toward this goal. 
}

\subsection{Benchmarking Algorithms on BASALT}
\label{sec:benchmark_code}
To assist with the development of LfHF algorithms and automated evaluations, we implement and share a codebase with two major contributions.
First, the code contains an example of training a LfHF algorithm with the shared data in the \demodataset.
Second, we include the tools for performing the evaluations presented in this work. 
The code is a Python-installable library, which allows the functionality to be imported into other codebases for use in research.

The training example provides tools to train a behavior cloning model on top of the Video PreTraining (VPT)~\citep{baker2022video} model using the \texttt{imitation}~\citep{gleave2022imitation} library.
VPT is a large foundation model that can complete various tasks in Minecraft, but it is difficult to fine-tune on new tasks due to its size. %
Inspired by the success of an imitation-learning approach\footnote{\url{https://github.com/shuishida/minerl_2022}} in the BASALT 2022 competition, we use behavior cloning as the base algorithm with the rich embeddings from the VPT model as input. %
Motivated by the original VPT results, we remove the no-op actions from the demonstration dataset. 
We provide code to use different variants of the VPT model for benchmarking.
The final output is a set of videos to use in evaluations against the shared recordings of other agents in the \evaldataset.

Because setting up human evaluations is time-consuming, we share our platform for conducting human evaluations. 
The platform is served as a webpage backed by a simple Python-based webserver. 
The collected data is either stored locally in a SQLite database or remotely in a more scalable form.
The platform includes a flexible API to add or remove agents from the set of comparisons.
After a simple setup, researchers can point human evaluators to a specific URL to provide answers.  
We provide examples of the form that the human evaluators will see in  \Cref{sec:eval_appx}. 
We also share code to create figures of the TrueSkill ratings (and all analyses in~\Cref{sec:analysis}), given the answers contained in the resulting database. This whole process serves as an end-to-end example of creating a new method with \demodataset, then evaluating it with data from \evaldataset.

\section{BASALT Evaluation and Demonstrations Dataset}
\label{sec:basalt_dataset}
We now introduce BEDD, our extensive dataset of human demonstrations and algorithm evaluations. 
This dataset consists of the following components: 
\change{
\begin{itemize}
    \item The \demodataset, a set of 13,928 videos (state-action pairs) demonstrating largely successful task completion attempts of the reward-free tasks, 
    \item The \evaldataset, a set of 3,049 dense pairwise comparisons of algorithmic and human agents attempting to complete the BASALT tasks, and
    \item The code for utilizing and analyzing these datasets for developing LfHF algorithms (some details in \Cref{sec:benchmark_code}).
\end{itemize}
}
For a full datasheet~\citep{gebru2021datasheets}, please see~\Cref{appx:datasheet}.

\subsection{\demodataset: Completing Reward-Free Tasks in Minecraft}
\begin{table}[t]
    \centering
    \begin{tabular}{lcccccc}
    \toprule 
     Task  & Videos & Episodes & Hours & Size & Ep. len, s& Success \% \\
     \midrule
     \cavetask  &  5,466 & 5,466 & 91 & 165GB & 60 & 93\% \\
     \waterfalltask & 4,230 & 4,176 & 97 & 175GB & 84 & 98\% \\ 
     \pentaskfull & 2,833 & 2,708 & 89 & 165GB & 119 & 95\% \\
     \housetaskfull & 1,399 & 778 & 85 & 146GB & 391 & 92\% \\
     \midrule 
     Total & 13,928 & 13,128 & 361 & 651GB & 99 & 95\% \\
    \bottomrule 
    \end{tabular}
    \caption{High-level demonstration data statistics decomposed by task. Episode length is the average episode length in seconds. A demonstration is counted as success if the player manually ended the episode instead of dying or timing-out.}
    \label{tab:demo_breakdown}
\end{table}

The \demodataset{} for developing new methods consists %
of 361 hours (26 million image-action pairs) of human demonstrations of the reward-free BASALT tasks.
This data consists of labeled trajectories, both with high-resolution image observations and keyboard and mouse actions for each frame.
In total, this is 651 GB of data. 
\Cref{tab:demo_breakdown} decomposes the high-level data statistics by task.
More details about this dataset are in \Cref{sec:demo_appx}.

Each demonstration consists of a trajectory $\tau = [s_0, a_0, \dots, s_N, a_N]$, or a sequence of state-action pairs, where $N$ is the trajectory length.
These pairs are contiguously sampled at every Minecraft game tick (20Hz).
Each state consists of %
the 640x360 RGB frame from the perspective of the player (see \Cref{fig:MC-screenshots}). %
Each action consists of two parts $a = [K, M]$, where $K$ is all keyboard interactions, $M$ is all ``mouse'' interactions (change in view, pitch, and yaw), mimicking the native human control interface of Minecraft.
This dataset serves as a starting point for using \textit{demonstrations} as a form of feedback to train agents.

\subsection{\evaldataset: Evaluating BASALT Agents} 
\begin{table}[t]
    \centering
    \begin{tabular}{lccccccccc}
    \toprule 
       &  &  & {Words in} &\multicolumn{3}{c}{Response Sentiment} \\
    Task & Comparisons & Hours & Response & \faThumbsOUp  & \faThumbsODown  & \rotatebox[origin=c]{270}{\faThumbsOUp} \\
     \midrule 
    \cavetask   & 722 & 60 & 27,948 & {79\%} & 14\% & 7\% %
    \\
    \waterfalltask & 682 & 56 & 26,437 & 76\% & 7\% & 17\% %
    \\
    \pentaskfull & 914 & 81 & 32,768 & 57\% & 11\% & 32\% %
    \\ 
    \housetaskfull & 731 & 76 & 26,917 & 63\% & 9\% &28\% %
    & \\ 
    \midrule 
    Total & 3,049 & 273 & 114,070\\
    \bottomrule 
    \end{tabular}
    \caption{High-level evaluation data statistics decomposed by task. 
    We report the total number of agent-agent comparisons, human labor hours, and words used in the natural-language justifications of selecting a specific agent as the best one. 
    We also report the percent of positive, neutral, and negative sentiments in these justifications.}
    \label{tab:evaldata_stats}
\end{table}

The \evaldataset{} contains 3,049 pairwise comparisons of different algorithms, produced from 273 hours of human labeling effort by 65 unique MTurk workers.
All responses are contained in a single JSON file. 
\Cref{tab:evaldata_stats} decomposes this dataset by task. 
Each evaluation in the dataset consists of the following:
(i) the names of the two agents used in the comparison, (ii) the corresponding videos shown to the human judge, (iii) an answer to the question of which player is better overall (Left, Right, Draw), (iv) a natural-language justification of this choice, (v) answers to at least one direct question about concrete achievements by the players (Left, Right, or Both), and answers to 4-7 comparative questions, such as which agent was more human-like (Left, Right, Draw, N/A).
The direct and comparative questions are task-specific.
Including the choice of which algorithm performed best, this dataset consists of a total of 27,905 comparisons along various factors.

We provide all responses in our public release of the data.
\change{We also} include a list of anonymized MTurk workers whose responses we found not to suit our standards (e.g., \change{providing the} same answer to every task). 
\change{Before performing the analyses reported in this paper, we filtered the data to exclude these responses.}
\smchange{We provide all details needed to understand and reproduce these evaluations in~\Cref{sec:eval_appx}.}
With this dataset, one may compare a newly-developed algorithm with $17$ possible agents: the top $13$ teams in the 2022 BASALT competition, a behavioral cloning baseline, a random agent, and two human experts (two of the authors of this paper). Researchers can use the provided human evaluations to kickstart the human evaluation, without needing to dedicate costly human labor to evaluating all algorithms from scratch.

\section{\change{Analysis:} \demodataset}
\label{sec:analysis}

\change{
We now analyze the \demodataset.
Because the BASALT tasks lack concrete reward functions, evaluating the progress of agents on these tasks is challenging. 
This difficulty necessitates the use of informative proxy measures. 
As a result, when analyzing the \demodataset, we focus on defining proxy measures that may be useful for understanding the data or tracking training progress.
In this section, we describe these proxies and present the results of our analysis.}

\change{We first seek to understand the relative difficulty of the four reward-free tasks.
Given the experience of the contracted data collectors with Minecraft, we believe that the length of the demonstration is a reasonable proxy for task difficulty.}
Using demonstration length as a proxy for task difficulty, we note that \housetaskfull is likely the most challenging task, even for humans: each video lasts around $6.52$ minutes on average, while the next most time-consuming task, \pentaskfullnospace, takes an average of around $1.98$ minutes (\Cref{tab:demo_breakdown}).
By this metric, the easiest task is \cavetask (1 minute).
In practice, one could use this metric to assess training progress or likelihood of task completion.
If an agent completes a task at a rate that is far away from the average, that may signal worse quality behavior.
In contrast, if an agent takes a similar amount of time to the average to complete a task, then that could signal better behavior (but is clearly not definitive).

\begin{figure}[t]
  \centering
  \begin{subfigure}[b]{0.49\textwidth}
    \includegraphics[width=\textwidth]{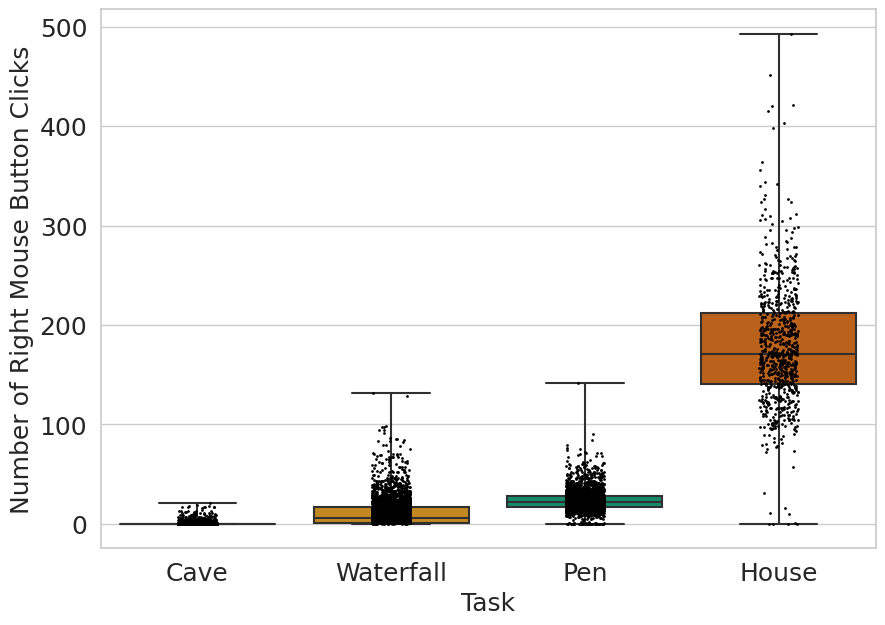}
    \caption{Right mouse button clicks}
  \end{subfigure}
  \begin{subfigure}[b]{0.49\textwidth}
    \includegraphics[width=\textwidth]{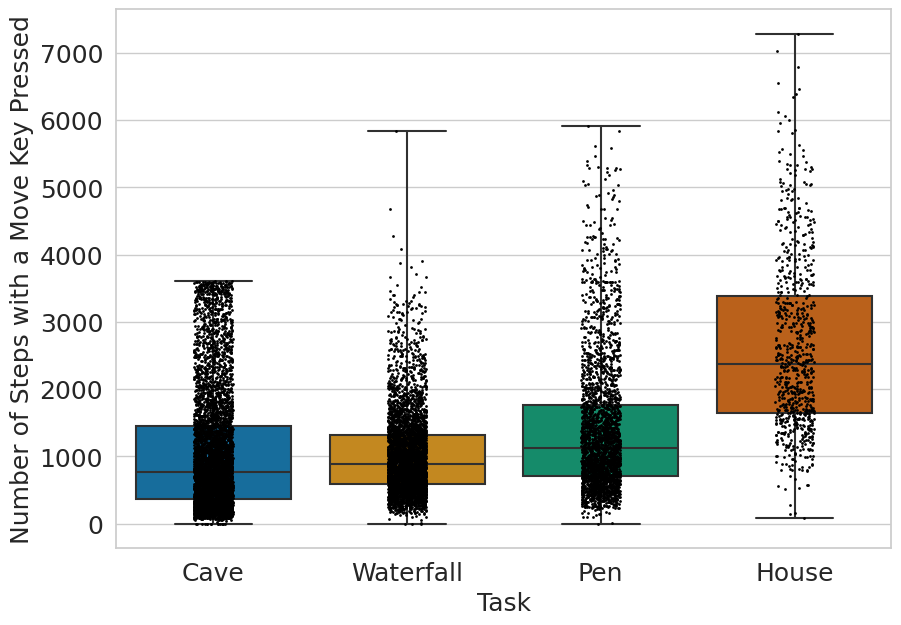}
    \caption{Movement key presses}
  \end{subfigure} \\
  \caption{The distributions of the number of right mouse button clicks and movement key presses across tasks in the \demodataset, which act as proxies for the number of blocks placed and distance traveled per episode, respectively. Because \housetaskfull takes the longest to complete, it is reasonable that it requires more right mouse button clicks and movement key presses.}
  \label{fig:RMB-clicks-and-movement-key-presses}
\end{figure}

\change{We also want to understand when an agent --- human or AI --- may be underperforming on the task.}
\change{Because \cavetask requires navigation to find a cave, an agent that remains stationary is likely unsuccessful.}
\change{Similarly, because \pentaskfull and \housetaskfull require the construction of objects, agents that fail to place any blocks likely do not succeed at the task.}
\change{As a result,} we employ the number of steps with an active movement key as a proxy for the distance traveled within the tasks.
We \change{also} use the count of right mouse button clicks as a proxy for the number of blocks placed.
The only other actions performed by clicking the right mouse button are using a crafting table or a chest, of which there are relatively few per episode.

These proxies are depicted in the per-episode distributions within the \demodataset{} in \Cref{fig:RMB-clicks-and-movement-key-presses}.
The number of right mouse button clicks is highest at around 180 in \housetask and between 20-40 in \waterfalltask and \pentaskfullnospace.
These align with expectations, given the amount of building required to complete the tasks.
The number of movement actions has the highest variability in \cavetasknospace.
This task also has the shortest time-out at 3,600 steps.
Perhaps a better proxy would be the average number of movement actions per step: \cavetask should have the highest number, as the task mainly consists of moving around.

\change{We include these results both as a way to understand this dataset and as values to monitor during agent training to estimate agent performance.}
However, it is crucial to avoid using these metrics as direct optimization targets, as they can be easily exploited.

\section{\change{Analysis:} \evaldataset} 
\label{sec:analysis_evaluation}

\smchange{
We first analyze the overall dataset, then focus on a few main comparisons: the top two algorithms from the BASALT 2022 competition (GoUp and UniTeam), the behavioral cloning baseline (BC-Baseline), the two human experts (Human1 and Human2), and a random agent (Random).}
\smchange{This subset of data corresponds to 394 of 3,049 total comparisons or nearly 34 hours of human labeling effort.
For additional details about the subsequent analyses, please see~\Cref{appx:eval_dataset_analyses}.}

\subsection{\change{Task-Based Analysis}}
\Cref{tab:evaldata_stats} provides a high-level overview of the 3,049 total evaluations, decomposed by task. 
The human judges generally responded using a similar number of words across the different tasks.
Only the responses to \cavetask and \waterfalltask contained significantly more words on average than \pentaskfull ($F=3.72$, $p=0.01$; 95\% CI: $0.34$ to $5.37$, $p=0.02$; 95\% CI: $0.36$ to $5.47$, $p=0.02$, respectively).\footnote{We conducted a one-way ANOVA with a post-hoc Tukey's HSD test to obtain these results. Results presented as (ANOVA $F$ value, $p$ value; CI for Tukey's HSD, $p$ value).\label{footnote:anova}} 
All other tasks exhibited no significant differences between means. 
This result suggests that expressing the specific rationale for these simpler tasks may be easier. 

\change{To understand the general perceptions of the human judges of the different tasks,} we categorized each response into positive, neutral, or negative sentiment~\citep{loria2018textblob}.
We then analyzed the differences in the distribution across tasks.
The human judges displayed different levels of sentiment in their responses, depending on the task.\footnote{We conducted a chi-square test of independence to examine the relationship between task and sentiment classification. 
The relation between these variables was significant, $X^2$ ($6, N =3049) = 132.21, p < .001.$} 
They responded most positively to \cavetask and least positively to \pentaskfullnospace. 
The only sentiment distributions that were \textit{not} significant were between \pentask and \housetaskfull ($X^2 (1, N=1645)= 6.35, p=0.042$), and \cavetask and \waterfalltask ($X^2 (1, N=1404) = 2.694, p=0.260$). All other pairs of tasks exhibited significant differences in sentiment.\footnote{These results are from Bonferroni-corrected pairwise Chi-square tests.\label{footnote:chi2_bonf}}
Due to the large number of subtask dependencies required for completing \pentaskfullnospace{} and \texttt{BuildVillageHouse} (as well as the greater amount of time required on average for human demonstrators to complete these two tasks, detailed in~\Cref{tab:demo_breakdown} and  \Cref{sec:analysis}), they are considered to be more challenging than \cavetask and \texttt{MakeWaterfall}. 
\change{The more positive sentiment toward \cavetask and \texttt{MakeWaterfall} may be due to an increased focus on successes due to their relative ease of completion.}

\subsection{\change{Agent-Based Analysis}}
\smchange{We now turn our attention to agent-based analysis.
We first provide a brief overview of the included agents.
GoUp uses human knowledge to decompose the tasks into the same high-level sequence, then uses computer vision techniques to identify the goal for each task (e.g., the cave).}
\smchange{UniTeam combines behavioral cloning with search by embedding the current set of images with a pre-trained VPT network and then searching for the nearest embedding point in the VPT latent space to find the situation to use as reference and copying the corresponding expert actions}~\citep{malato2022behavioral}.
More details about the other assessed agents can be found in previous work~\cite{milani2023towards}.

\begin{table}[t]
    \centering
    \begin{tabular}{lcccccc}
    \toprule 
     & Normalized &\multicolumn{3}{c}{Sentiment} \\
    Agent & TrueSkill  & \faThumbsOUp  & \faThumbsODown  & \rotatebox[origin=c]{270}{\faThumbsOUp} \\ 
    \midrule 
    Human2   & $2.43$ & $92$\% & $5$\% & $3$\% \\ 
    Human1 & $2.17$ &$92$\% & $6$\% & $2$\% \\
    GoUp & $0.73$ & $74\%$ & $19\%$ & $7\%$\\ 
    UniTeam & $0.13$ & $66\%$ & $25\%$ & $9\%$\\
    BC-Baseline & $-0.32 $ & $65\%$ & $26$\% & $9$\% \\
    Random & $-1.35$  &$63$\% & $29$\%  & $8$\% \\
\bottomrule
    \end{tabular}
    \caption{Normalized TrueSkill score and percent of positive, neutral, and negative sentiments of the natural language justifications for selecting an agent as the best one.}
    \label{tab:eval_agents}
\end{table}

We present the agent-based overview of the evaluations in 
\Cref{tab:eval_agents}.
Note that the reported TrueSkill scores are computed over the 3,049 total comparisons, whereas the sentiment analysis is performed only on the 394 agent-specific entries. 
The human agents performed the best \change{using TrueSkill ranking}, with a large performance gap between them and the best-performing algorithmic agent. 
This result emphasizes the difficulty of BASALT for algorithmic agents.

The only sentiment distributions that \textit{were} significant were between either of the two human experts and the other algorithms. 
In particular, the sentiment toward comparisons that included human players were significantly more positive.\footref{footnote:chi2_bonf} 
All other pairs of algorithms did not exhibit significant differences in sentiment.
In general, human judges were generally positive when evaluating the agents. The judges may have been more positive due to social influence: they knew that the videos were produced by different players for ranking in the competition.

\begin{figure}[t]
  \centering
  \begin{subfigure}[b]{0.495\textwidth}
    \includegraphics[width=\textwidth]{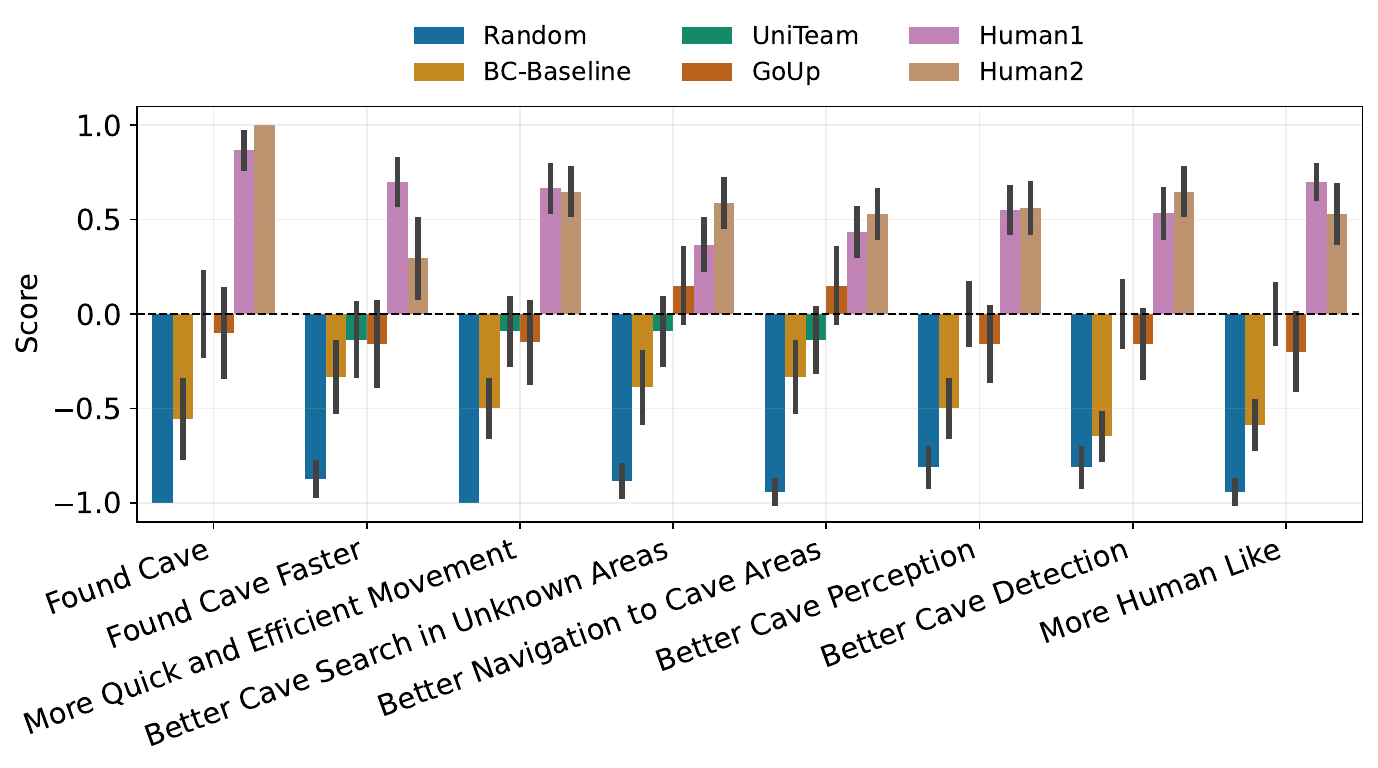}
    \caption{\cavetasknospace}
    \label{fig:cave_human_eval_questions}
  \end{subfigure}
  \begin{subfigure}[b]{0.485\textwidth}
    \includegraphics[width=\textwidth]{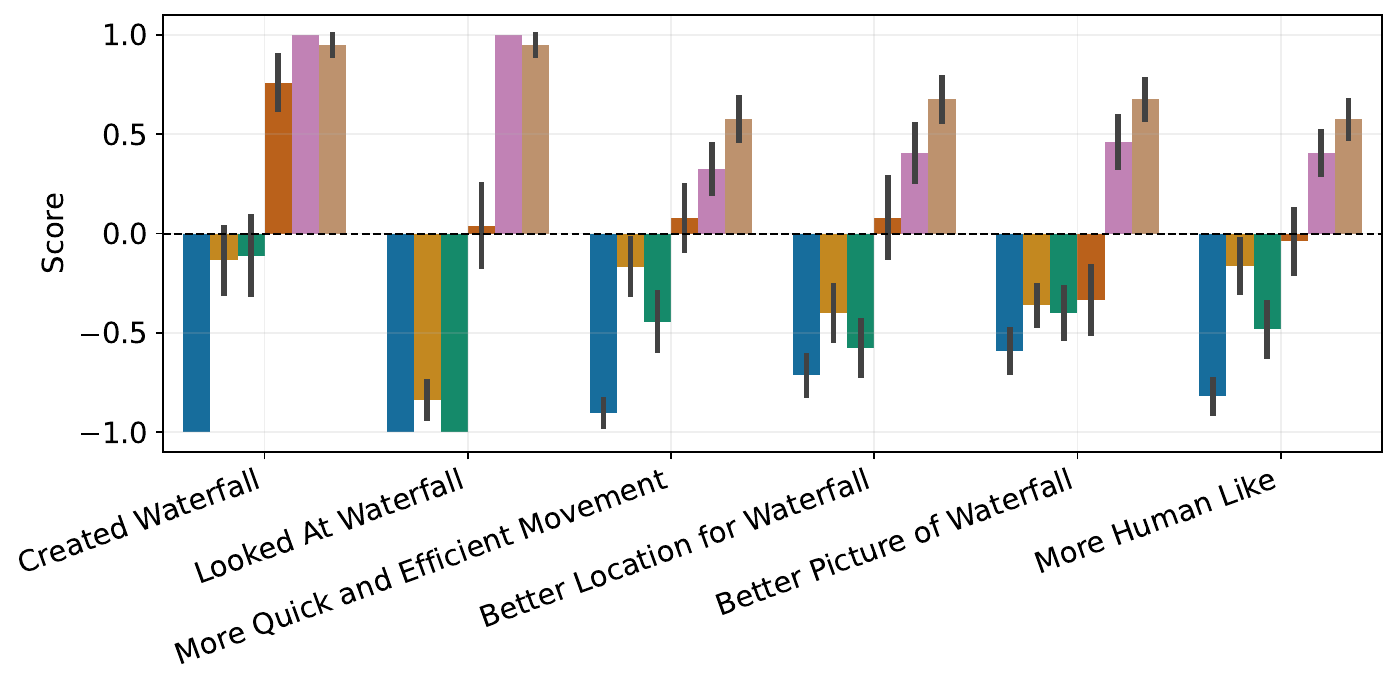}
    \caption{\waterfalltask}
    \label{fig:waterfall_human_eval_questions}
  \end{subfigure} \\
  \begin{subfigure}[b]{0.49\textwidth}
    \includegraphics[width=\textwidth]{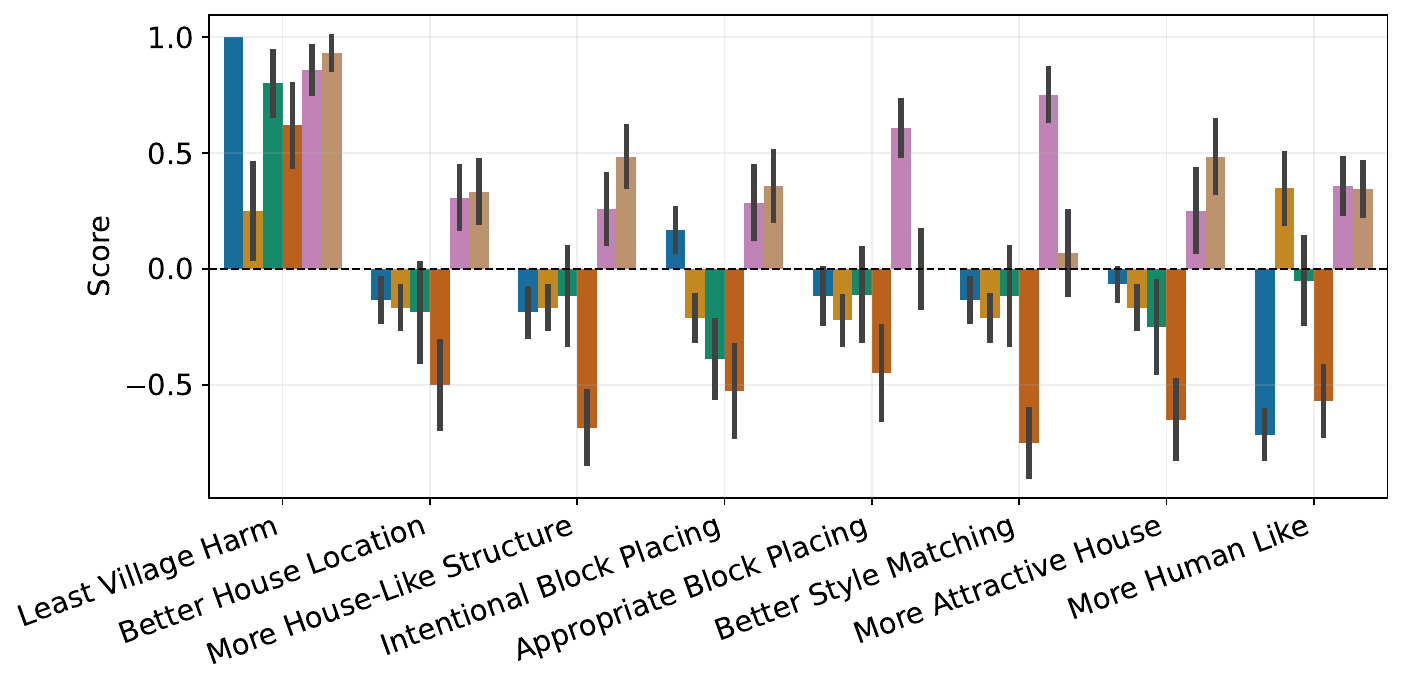}
    \caption{\housetaskfull}
  \end{subfigure}
  \begin{subfigure}[b]{0.49\textwidth}
    \includegraphics[width=\textwidth]{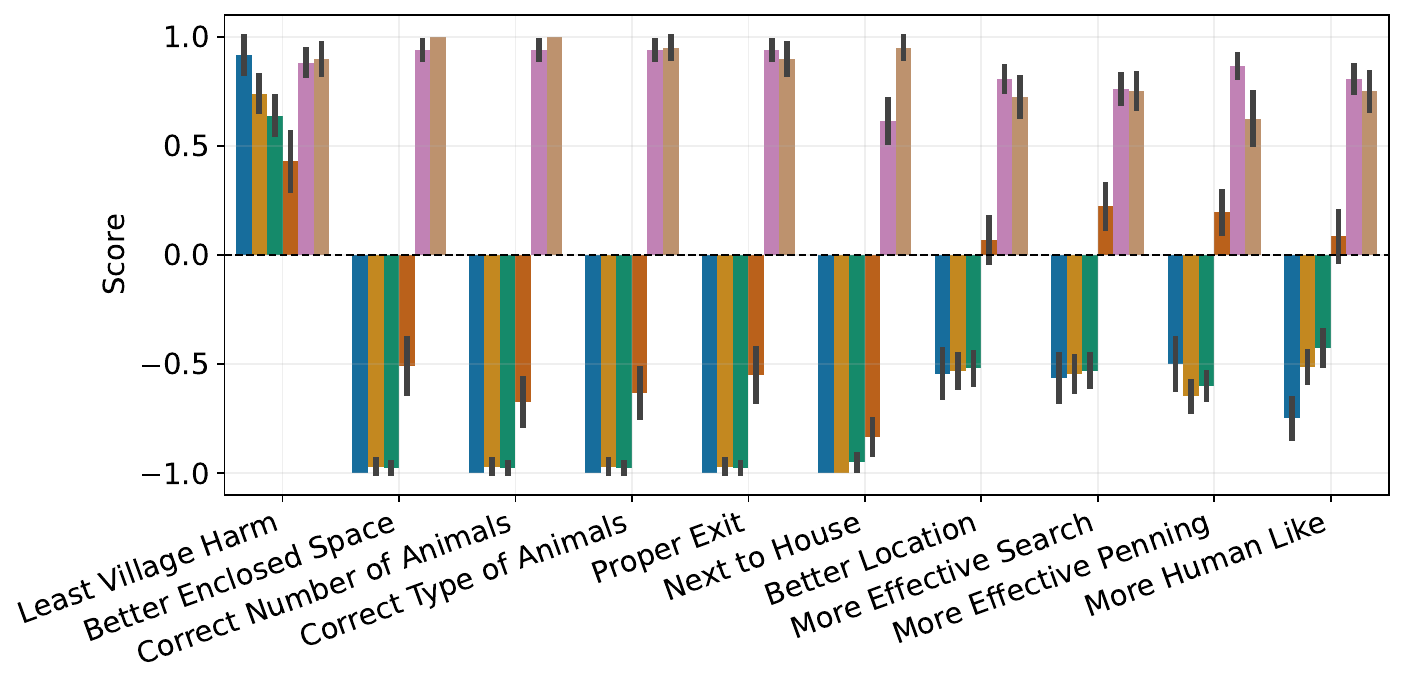}
    \caption{\pentasknospace}
  \end{subfigure}
  \caption{Comparison of baseline solutions, top BASALT 2022 competition solutions, and humans on additional questions. Bars represent average score, and error bars represent standard error. Higher is better: it means that agent exhibited more of that factor according to human judges.} %
  \label{fig:human_eval_questions}
\end{figure}

\smchange{
We now investigate the responses to the direct and comparative questions. 
For succinctness, we summarize each question into a single factor.
For example, we use ``More Human Like'' to capture the question, ``Which player seemed more human-like (rather than a bot or computer player)?''.
\Cref{fig:human_eval_questions} presents these results.
Importantly, no machine learning algorithms outperform humans on any of the factors, indicating that there is still plenty of improvement to be made.
When analyzing future algorithms, we suggest preserving this decomposition to both validate that the responses are sensible and obtain a fine-grained understanding of agent capabilities.
To validate sensible responses, we suggest checking the scores for the \textit{random} agent: except for causing the least harm, this agent should score poorly compared to the other agents.
}

\change{Finally, we highlight a few interesting findings that are enabled by our decomposed and extensive set of evaluation criteria.
\Cref{fig:waterfall_human_eval_questions} (\texttt{MakeWaterfall}) in the paper reveals that, although Team GoUp’s algorithm can create waterfalls at a rate more similar to the human players, it struggles along all other criteria, including choosing a good location and taking a high-quality photograph. 
Only looking at the binary success/failure condition for creating a waterfall would ignore these important nuances in the algorithm’s behavior. 
As another example, \Cref{fig:cave_human_eval_questions} (\texttt{FindCave}) suggests that, while Team GoUp’s algorithm still struggles to find caves, it can reasonably search for and navigate to areas that are likely to have caves. 
This finding suggests that the performance bottleneck may be the cave detection system employed by this approach. 
}

\section{Related Work}
\label{sec:related-work}

\paragraph{Learning from Human Feedback}

Learning from human feedback has become a crucial research direction in machine learning \cite{arumugam2019deep,christiano2017deep,shah2021minerl}, aiming to leverage human expertise to improve algorithm performance and generalization. 
Various algorithms have been proposed to integrate human feedback into the learning process, often using techniques such as imitation learning \cite{hoque2021thriftydagger}, inverse reinforcement learning \cite{brown2018risk, wu2023human}, and reward modeling \cite{stiennon2020learning}. 
Most commonly, these algorithms are evaluated in standard reinforcement learning benchmarks \cite{ bellemare2013arcade, cobbe2019leveraging, 
duan2016benchmarking,tassa2018deepmind}. 
However, most of these benchmarks do not have the property that human feedback is \textit{critical} for task identification, as in BASALT.
In some Atari games, if an agent does anything other than the intended gameplay, it dies and resets to the initial state, so pure curiosity-based agents perform well \cite{burda2018large}.
In contrast, BASALT tasks require human feedback or data to identify and complete tasks.

\paragraph{Minecraft for Machine Learning}

Minecraft is as a valuable platform for machine learning research~\citep{hafner2023mastering,lifshitz2023steve,smit2023game} due to its complex, dynamic environment and customizable game engine. 
As a result, numerous competitions and benchmarks~\citep{grbic2021evocraft,hafner2021benchmarking,johnson2016malmo,salge2018generative} have been developed to improve AI capabilities. %
These often include tasks with concrete or programmatic reward functions, such as learning to collect diamonds in a sample-efficient way~\citep{guss2019neurips}, multi-agent learning of cooperative and competitive tasks~\citep{perez2019multi}, and more~\citep{fan2022minedojo}.
In contrast, we emphasize tasks that are designed to be hard to specify through a reward signal.
Other work focuses on natural language as a specific modality for communicating intent~\citep{gray2019craftassist, kiseleva2021neurips,kiseleva2022iglu, narayan-chen-etal-2019-collaborative}; instead, we aim to promote the development of algorithms that generally learn from human feedback, including natural language. %
Some techniques require complete game state information (knowledge of blocks and items around them) \cite{wang2023voyager,zhu2023ghost}; however, our benchmark only permits agents to access observations, not game state, to promote learning from pixels, which is more generalizable to other tasks.

\paragraph{Comparison to MineDojo} Perhaps the most similar benchmark to ours is MineDojo~\citep{fan2022minedojo}, which emerged during the BASALT competitions and contributed a large dataset of Minecraft gameplay with the aim of developing generally-capable embodied agents. 
\change{The focus of MineDojo was to provide a massive dataset scraped from the internet; in contrast, we focus on curating a smaller set of high-quality demonstrations and evaluations. 
Our demonstration data was produced by experienced Minecraft data collectors using a consistent Minecraft version and settings.}
\change{In contrast, the MineDojo data, although plentiful, contains many videos with streamer overlays and non-Minecraft parts, different texture packs, mod packs, and more. 
Although this diversity may be beneficial in some settings, previous work had to filter out such data before it could be useful for training~\citep{baker2022video}.}
\change{Another critical difference is evaluation: MineDojo provides only} binary success or failure criterion; instead, our \change{recommended evaluation and resulting} \evaldataset{} \change{involves human judgments} across a range of quantitative (e.g., Found Cave) and qualitative (e.g., Style Matching) criteria.

\section{Conclusion}
\label{sec:conclusion}
\smchange{
We proposed \fulldataset, a large and accessible dataset to facilitate algorithm development for the BASALT benchmark on learning from human feedback.
The benchmark consists of five tasks in Minecraft, four of which lack a reward function. 
We clarified the benchmark, providing concrete evaluation recommendations and introducing another avenue for benchmarking: automated evaluations.
Our dataset, \fulldataset, consists of two key parts: \demodataset{} and \evaldataset.
\demodataset{} contains almost 14,000 videos showing successful task completions of the four reward-free tasks.
The \evaldataset{} consists of over 3,000 human evaluations to support the development of automated evaluations of hard-to-specify tasks. 
To our knowledge, this dataset is the largest of its type: one that supports both training and evaluating LfHF agents on tasks with hard-to-specify reward functions.
}%

\smchange{
We demonstrated the utility of our dataset and benchmark by presenting an analysis of the demonstrations and the several algorithms in the \evaldataset. 
We showed that the algorithms exhibit varying degrees of performance on our tasks, as evaluated by human judges on a variety of factors.
Our results suggest that there is ample room for improvement in learning from human feedback.
With the inclusion of our accessible code for benchmarking and evaluating agents, we hope that our contributions will encourage the development of more effective approaches for both learning from human feedback and evaluating these techniques in the future.
}

\ack
Creating these datasets and this benchmark was only possible with the help of many people and organizations.
FTX Future Fund Regranting Program, Microsoft, Encultured AI, and AI Journal provided financial support for the competition and resulting data. 
We thank Berkeley Existential Risk Initiative (BERI) for the support in organizing the BASALT competition.
Karolis Ramanauskas was supported by the UKRI Centre for Doctoral Training in Accountable, Responsible and Transparent AI (ART-AI) [EP/S023437/1] and the University of Bath.
We thank our amazing advisory board of the 2022 competition --- Fei Fang, Kiant\'e Brantley, Andrew Critch, Sam Devlin, and Oriol Vinyals --- for their advice and guidance. 
We also thank any previous organizers or advisors of the previous competitions for their contributions.
Finally, we thank AIcrowd for their help hosting the competion and the MTurk workers for their work evaluating the submissions.

\bibliography{bib}

\newpage

\newpage
\appendix

\section{BASALT Benchmark}
\label{sec:benchmark_appx}

\subsection{Evaluation Protocol}
Here, we reiterate the rules previously specified in \Cref{sec:basalt}:
\begin{itemize}
    \item Only pixel observations provided by the four environments should be used. You may read other observations from the environment for developing purposes, but the agent (or the system acting) should \textit{only} see the image pixel observations.
    \item The environments should not be modified in any way.
    \item Action shaping is okay, as long as it is done in line with other methods.
    \item Algorithms should be evaluated with TrueSkill using the specific hold-out environment test seeds provided below.
    \item The final evaluations must be conducted with human evaluators.
\end{itemize}

Evaluation environment seeds for each task:
\begin{itemize}
    \item \texttt{MineRLBasaltFindCave-v0}: 14169, 65101, 78472, 76379, 39802, 95099, 63686, 49077, 77533, 31703, 73365.
    \item \texttt{MineRLBasaltMakeWaterfall-v0}: 95674, 39036, 70373, 84685, 91255, 56595, 53737, 12095, 86455, 19570, 40250.
    \item \texttt{MineRLBasaltCreateVillageAnimalPen-v0}: 21212, 85236, 14975, 57764, 56029, 65215, 83805, 35884, 27406, 5681265, 20848.
    \item \texttt{MineRLBasaltBuildVillageHouse-v0}: 52216, 29342, 67640, 73169, 86898, 70333, 12658, 99066, 92974, 32150, 78702.
\end{itemize}

\subsection{TrueSkill Procedure}
\label{app:trueskill_procedure}
Please see the previous specification \citep{milani2023towards} and the evaluation server code.

\subsection{Discussion of Evaluation Metrics}
\label{app:eval_metric_options}
\begin{table}[t]
    \centering
    \begin{tabular}{lccc}
    \toprule 
         &  TrueSkill & Elo & Individual Analysis \\
    \midrule 
    Easily compare algorithms & \cmark & \cmark & \xmark \\
    Handle multi-algorithm comparisons & \cmark & \xmark & \xmark \\ 
    Provide uncertainty estimates & \cmark & \xmark & \xmark \\
    \bottomrule
    \end{tabular}
    \caption{Comparison of different potential evaluation systems for the BASALT benchmark.}
    \label{tab:evaluation_comparison}
\end{table}
\smchange{
We choose TrueSkill as the metric for assessing the performance of agents on the BASALT task.
In this section, we provide a detailed comparison of TrueSkill with other alternatives.
For a summary of these comparisons, see~\Cref{tab:evaluation_comparison}.
}

\paragraph{TrueSkill vs. Elo}
\smchange{
The Elo rating system~\citep{elo1978rating} is a popular zero-sum system that was initially created for chess. 
The gain in rating points of a player, or algorithm, after a win is equivalent to the loss in points for their competitor. 
The outcome of the game and the difference in ratings between algorithms influence the specific number of points transferred. 
Elo was designed for two-player games, whereas TrueSkill can handle multi-player games.
In the case of BASALT, this means that human evaluations can be performed over more than two algorithms, which may be more cost-effective for researchers to deploy.
Furthermore, Elo does not incorporate uncertainty estimates for an algorithm's performance, whereas TrueSkill does.
This uncertainty estimate not only makes results easier to interpret but also enables more adaptive ratings: performance estimates more quickly adjust for new algorithms because of high uncertainty and less quickly adjust for more established algorithms because of lower uncertainty.
}

\paragraph{TrueSkill vs. Independent Algorithm Performance Assessment}
\smchange{
Due to the difficulty of the BASALT tasks, independently assessing the performance of an algorithm is challenging.
For example, in the absence of any comparisons, how would one assess the performance of an algorithm that only partially completes a task? 
How could we ensure that the score is calibrated for the assessment of other algorithms? 
These standalone assessments do not offer ways to directly compare multiple algorithms or adjust the rating of an algorithm based on its performance over time, as TrueSkill and Elo do.
}

\subsection{Task Descriptions}
\label{appx:task_descriptions}

Here, we provide more detailed descriptions of the four reward-free BASALT tasks.

\paragraph{BuildVillageHouse} \smchange{The agent spawns in a village, which is a naturally-occurring structure in Minecraft that contains houses and other buildings, villagers, and other structures. 
The style of the buildings depends on the biome in which the village is generated.
The agent must build a new house for itself in the same style as the other village houses; however, it must not damage the village.
To assist agents with achieving this task, we provide the agent with the materials required to build a house in a variety of village types.
}

\paragraph{CreateVillageAnimalPen}
\smchange{
The agent spawns in a village. 
It is tasked with building an animal pen next to one of the houses in the village.
It must then corral a pair of farm animals (chickens, cows, sheep, or pigs) into the pen. 
In Minecraft, players perform this task to breed the animals to create a steady source of food.
We provide the agent with materials to use to build the pen.
Because different animals are lured with different types of food, we also equip the agent with the food to lure any of the animals.}

\paragraph{MakeWaterfall}
\smchange{The agent must create a waterfall and subsequently take a picture of it.
The agent spawns in an extreme hills biome to enable easier task completion. 
Because taking a photograph is not a supported task in Minecraft, we implement it by having the agent throw a snowball.
We interpret the moment that the agent throws it as its photograph. 
The challenge with this task is it involves both the evaluation of task completion and human aesthetic preferences. 
We provide the agent with the tools needed for efficiently moving around the often-treacherous biome, a snowball for taking a picture, and two buckets of water to construct the waterfall. }

\paragraph{FindCave}
\smchange{The objective of this task is for the agent to discover a naturally-generated cave.
The agent spawns in a plains biome and must explore the surrounding area to find a cave.}
\section{Datasheet for \fulldataset{}}
\label{appx:datasheet}

We present a datasheet \citep{gebru2021datasheets_appendix} for \fulldataset{}, which includes details on our 651GB human demonstrations dataset (\demodataset) and our TrueSkill evaluations dataset (\evaldataset).

\subsection{Motivation}

\textbf{For what purpose was the dataset created?}

Originally, both datasets were created for the BASALT 2022 competition \cite{milani2023towards}. The \demodataset{} was provided as a training dataset for competitors, while the \evaldataset{} was generated at the end of the competition when evaluating submissions. Below, we describe further purposes for the datasets.

\paragraph{\demodataset}
\smchange{
We believe that this set of successful demonstrations could be useful for training machine learning algorithms to perform complex tasks in Minecraft.}
\smchange{
All task completions can be thought of as having a positive label denoting successful completion.} 
\smchange{The dataset could serve as a basis for generating pretrained embeddings, similar to the VPT method.
These embeddings could be further finetuned with additionally collected demonstrations, corrections, or other feedback modalities to further refine the behavior.}
\smchange{
This dataset could also be leveraged for a wealth of insights.}
\smchange{
Since these tasks are complex and require sequential dependencies, one could analyze and extract the crucial subtasks required to complete the full task. 
This analysis would support the development of various hierarchical methods.}
\smchange{This task decomposition would support further subtask-specific analysis.
Understanding the required time to complete each subtask would help identify which tasks take longer or are completed more quickly.
This information could provide meaningful insights into task difficulty. }

\paragraph{\evaldataset}
\smchange{
We envision this dataset to be used for both analysis and development.
The detailed task-specific questions may be useful for understanding which factors most strongly influence overall assessments of behavior.}
\smchange{We envision this dataset to be used to develop reward models that better align with human preferences. 
This aim aligns with the BASALT benchmark of both understanding and utilizing human preferences to solve fuzzy tasks. 
Training reward models on pairwise comparisons over algorithms may reveal the latent structure of human preferences over the task space.
A more nuanced understanding would emerge as more of the task-specific questions and long-form justification of responses are processed and understood.
This training regime would promote the development of more intuitive and human-aligned algorithms.}\footnote{
We again stress that the goal is not necessarily to \textit{completely replace} the human element involved in the evaluation of these algorithms. 
Instead, automated evaluations serve as a way to sidestep the issues with existing evaluation methods, which involve laborious manual assessments or rely on predefined metrics that may fail to capture all aspects of an algorithm's performance. 
With a dataset of human-based comparisons, we can train machine learning models to predict the superior algorithm for a given task, thereby reducing the need for extensive manual evaluations.}
\smchange{Furthermore, we envision that this dataset could provide a wealth of insights into more general human preferences, which could prove invaluable when transferred or used to warm-start reward models in different domains. 
By identifying these underlying general preferences, we could inform the design of more universally acceptable and effective AI systems (bearing in mind individual, cultural, and other important differences).
As an example, people may favor algorithms that show a more structured, stepwise approach to problem-solving tasks. 
One interpretation of this preference is orderly progression, which could help design algorithms that behave in ways that are more interpretable. }

\textbf{Who created the dataset?} 

The dataset was created by Anssi Kanervisto (Microsoft Research), Stephanie Milani (Carnegie Mellon University), Karolis Ramanauskas (University of Bath), Byron V. Galbraith (Seva Inc.), Steven H. Wang (ETH Zürich), Sander Schulhoff (University of Maryland), Brandon Houghton (OpenAI), Sharada Mohanty (AIcrowd), and Rohin Shah. The dataset was not created on the behalf of any entity.

\textbf{Who funded the creation of the dataset?}

The FTX Future Fund Regranting Program, Encultured.ai, and Microsoft funded the creation of the \evaldataset, prizes, and compute. OpenAI sponsored the creation of the \demodataset.

\subsection{Composition}
\label{appx:Composition}

\textbf{What do the instances that comprise the dataset represent (e.g., documents, photos, people, countries)?}

The \demodataset{} contains videos of agent trajectories and associated actions taken. The \evaldataset{} contains videos of trajectories and human evaluations on multiple factors such as "Found Cave" or "Least Village Harm".

Please refer to \Cref{sec:basalt_dataset} for more information. \Cref{tab:demo_breakdown} contains a breakdown of information on the \demodataset{} and \Cref{tab:evaldata_stats} contains a breakdown of information on the \evaldataset. 
Further information is in \Cref{appx:demo_dataset_details,sec:eval_appx}.

\textbf{Is there a label or target associated with each instance?}

The \evaldataset{} contains labeled data. Agents are evaluated across a range of quantitative and qualitative criteria (\Cref{sec:analysis_evaluation}). Some of these, such as "Found Cave" (\Cref{appx:eval_dataset_analyses}) can be considered to be labels. Additionally, the \demodataset{} contains videos of trajectories paired with `labels' of the action(s) being executed at each frame. Our codebase contains files that describe how each trajectory ends (e.g., by ESC key or death.).

\textbf{Is any information missing from individual instances?} 

Not to our knowledge.

\textbf{Are there recommended data splits (e.g., training, development/validation, testing)?} 

No.

\textbf{Are there any errors, sources of noise, or redundancies in the dataset?} There are instances of contractors doing a task incorrectly in the \demodataset. For example, in some instances, contractors found a cave and did not stop the episode or made a waterfall and stared at it instead of ending the episode.
Please see \Cref{appx:demo_dataset_details} for more possible issues.
We did not manually check the entire dataset, so it may contain additional anomalous activities.

\textbf{Do/did we do any data cleaning on the dataset?}

We did not. All data is presented exactly as collected. We provide information on which demonstrations may contain human errors in the repository. 
This information may be then used to conduct data cleaning by the end user of the dataset.

\subsection{Collection Process}
\label{appx:Collection-Process}

\textbf{How was the data associated with each instance acquired?}

The \demodataset{} was collected using a modded, closed source Minecraft version that recorded users' game frames and actions. Contractors watched agent gameplay side by side in order to evaluate them and create the \evaldataset{}.

\textbf{Who was involved in the data collection process and how were they compensated?}

For the \demodataset{}, 20 contractors were hired and paid 20 USD an hour. Additionally, some members of the BASALT team collected data for this dataset. For the \evaldataset{}, 65 high quality crowdsource workers from Amazon Mechanical Turk were paid to collect the data at approximately 15 USD an hour.

\textbf{Over what timeframe was the data collected?}

The \demodataset \ was collected between April 2022 and July 2022. The \evaldataset \ was collected between November 2022 and January 2023.

\subsection{Uses}

\textbf{Has the dataset been used for any tasks already?}

Both datasets were used for the BASALT 2022 competition. Please see more details in the competition retrospective \cite{milani2023towards} and this competitor white paper \cite{malato2022behavioral}.

\textbf{Is there a repository that links to any or all papers or systems that use the dataset?}

This \href{https://minerl.readthedocs.io/en/latest/notes/useful-links.html}{page} contains a list of papers and project that use MineRL. Only some projects from 2022 onward use \fulldataset{} data. Competitor papers from the 2022 competition \cite{malato2022behavioral} only used the \demodataset{} (the \evaldataset{} was created when evaluating these submissions).

\textbf{Is there anything about the composition of the dataset or the way it was collected and preprocessed/cleaned/labeled that might impact future uses?}

We do not believe so because the data for both datasets was collected from paid contractors and high-quality paid crowdsourcers.

\subsection{Distribution}

\textbf{Will the dataset be distributed to third parties?}

Yes, it is free and available online.

\textbf{How will the dataset will be distributed (e.g., tarball on website,
API, GitHub)? Does the dataset have a digital object identifier (DOI)?}

The \evaldataset{} exists on 
Zenodo (DOI:
10.5281/zenodo.8021960) as a zip file.

The \demodataset{} exists on an OpenAI server as JSONL and MP4 files and does not have a DOI. 

All data is under the MIT license.

\textbf{Have any third parties imposed IP-based or other restrictions on the data associated with the instances?}

No.

\textbf{Do any export controls or other regulatory restrictions apply to the dataset or to individual instances?}

No.

\subsection{Maintenance}

\textbf{Who will be supporting/hosting/maintaining the dataset?}

The authors on this paper will provide needed maintenance to the datasets (\demodataset{}, \evaldataset{}). We do not expect much maintenance to be needed as we will not be adding data to the dataset. However, we will accept PRs from users who have improvements to make to the supporting codebase.

\textbf{How can the owner/curator/manager of the dataset be contacted
(e.g., email address)?}

Please email us at basalt@minerl.io

\textbf{Is there an erratum?}

There is not, but 1) we mention potential issues with the data in this datasheet, and 2) we provide a list of data within the dataset that we believe to be invalid due to issues such as not properly completing the given task (\Cref{appendix:idiosyncratic}). 

\textbf{Will the dataset be updated (e.g., to correct labeling errors, add
new instances, delete instances)?}

Yes, but we expect minimal updates to be required, as we do not intend to add more data to the dataset.
\section{Demonstrations Dataset Details}
\label{sec:demo_appx}

\subsection{Human Data Collection Details}
Contractors were originally hired through Upwork by responding to the following job posting:
\begin{formal}
We are looking for people who want to get paid to play Minecraft. We will want you to describe some things about your experience, so our ideal candidate would possess Native English fluency and have a microphone. You'll need to install java, download a modified version of Minecraft (that collects and uploads your play data and voice), and play Minecraft survival mode! Paid per hour of gameplay. Prior experience in Minecraft is not necessary. We do not collect any data that is unrelated to Minecraft from your computer.
\end{formal}

Prior to recording this dataset, all contractors had been working on other datasets in Minecraft\footnote{Previous Minecraft experience was not a hard requirement in the job posting. However, in practice, those without experience did not continue their work.}, and were proficient in taking English language instructions via UpWork and using the provided recording scripts for generating demonstration data. 
The contractors were initially hired in the context of collecting narrated Minecraft play.
However, narrations were not requested for the BASALT dataset, and any incidentally collected narrations will not be released.
In total, 20 contractors were recruited and paid $20$ USD per hour. 
We allocated a total of $10$k USD for this.
To supplement this dataset, three members of BASALT contributed data.
Below are the exact transcripts we used to instruct the contractors for the four tasks.

\subsubsection{\cavetask}
\begin{formal}
Task 1 - Recorder version find-cave
Look around for a cave. When you are inside one, quit the game by opening main menu and pressing "Save and Quit To Title".
You are not allowed to dig down from the surface to find a cave.

Timelimit: 3 minutes.

Example recordings: \url{https://www.youtube.com/watch?v=TclP_ozH-eg}
\end{formal}

\subsubsection{\waterfalltask}
\begin{formal}
After spawning in a mountainous area with a water bucket and various tools, build a beautiful waterfall and then reposition yourself to “take a scenic picture” of the same waterfall, and then quit the game by opening the menu and selecting ``Save and Quit to Title"

Timelimit: 5 minutes.

Example recordings: \url{https://youtu.be/NONcbS85NLA}

\end{formal}

\subsubsection{\pentaskfull}
\begin{formal}
After spawning in a village, build an animal pen next to one of the houses in a village. Use your fence posts to build one animal pen that contains at least two of the same animal. (You are only allowed to pen chickens, cows, pigs, sheep or rabbits.) There should be at least one gate that allows players to enter and exit easily. The animal pen should not contain more than one type of animal. (You may kill any extra types of animals that accidentally got into the pen.) Don’t harm the village. After you are done, quit the game by opening the menu and pressing "Save and Quit to Title".

You may need to terraform the area around a house to build a pen. When we say not to harm the village, examples include taking animals from existing pens, damaging existing houses or farms, and attacking villagers. Animal pens must have a single type of animal: pigs, cows, sheep, chicken or rabbits.

The food items can be used to lure in the animals: if you hold seeds in your hand, this attracts nearby chickens to you, for example.

Timelimit: 5 minutes.
Example recordings: \url{https://youtu.be/SLO7sep7BO8}

\end{formal}

\subsubsection{\housetaskfull}
\begin{formal}
Taking advantage of the items in your inventory, build a new house in the style of the village (random biome), in an appropriate location (e.g. next to the path through the village), without harming the village in the process. Then give a brief tour of the house (i.e. spin around slowly such that all of the walls and the roof are visible).

You start with a stone pickaxe and a stone axe, and various building blocks. It’s okay to 
break items that you misplaced (e.g. use the stone pickaxe to break cobblestone blocks).
You are allowed to craft new blocks.

Please spend less than ten minutes constructing your house.

You don’t need to copy another house in the village exactly (in fact, we’re more interested in
having slight deviations, while keeping the same ``style"). You may need to terraform the area
to make space for a new house. When we say not to harm the village, examples include taking animals from existing pens, damaging existing houses or farms, and attacking villagers.

After you are done, quit the game by opening the menu and pressing ``Save and Quit to Title".

Timelimit: 12 minutes.

Example recordings: \url{https://youtu.be/WeVqQN96V_g}

\end{formal}

\subsection{Dataset Details}
\label{appx:demo_dataset_details}

Like most other datasets, this one contains some issues.
Instead of waiting for users to discover them, we preemptively investigate the dataset ourselves.
We document the issues below.
This is done in the interest of transparency and to ensure the dataset is maximally useful.

\subsubsection{Episode Boundaries}
The recording software employed for our dataset splits video and action label files into 5-minute segments. This has no bearing on the \cavetask task, as its time limit is 3 minutes.
However, the time limits for \waterfalltask and \pentaskfull tasks are 5 minutes, and occasionally episodes exceed this limit by up to 3 seconds, triggering the video splitter. For the \housetask task, the time limit is 12 minutes, causing some episodes to be divided into three parts.

Certain training or analysis methods necessitate knowledge of episode boundaries, requiring a reliable method for identifying which videos belong to the same episode. 
While file labels being unique to an episode would simplify this process, this is not the case. We investigated various systematic approaches to detect where an episode ends and a new one begins, but none proved 100\% reliable. Some challenges we faced include: different episodes sharing the same file ID; video splitter not generating filenames with timestamps exactly 5 minutes apart; ESC keys not consistently triggering episode ends, because ESC is also used to close inventory; some episodes concluding in exactly 5 minutes; and a few episodes missing their first 5 minutes, rendering the loading screen in the initial video frame an unreliable boundary indicator. Our most reliable and straightforward solution was to consider two files as part of the same episode if the first file is exactly 5 minutes long, no ESC key is pressed with the GUI closed, and the filename ID is identical. This approach resulted in an error rate of less than 1\%, based on manual inspection of the first frames of all videos in the final split.

Leveraging this heuristic, we produced a file for each task containing a list of episodes, with each episode having an associated file list. We also included the step count per episode and two tags specifying (1) whether the episode ended with an ESC key press, (2) whether it was incomplete due to a saving error. Episode ending with an ESC key press indicates a successfully completed task. If an episode does not conclude with an ESC key press, the step count can be utilized to determine the type of episode end — a step count at or slightly above the time limit implies a timeout, while a lower step count indicates player death. The four files are located in the same repository mentioned above. This is the recommended way of using the dataset if episode boundaries are important for the training algorithm. 

\subsubsection{Idiosyncratic Episodes}
\label{appendix:idiosyncratic}
We also noticed some episodes, which pass our filters for being valid and complete episodes but have some unique characteristics. This might be useful to know for data cleaning purposes.
We provide some examples below, including the associated file names.

\begin{itemize}
    \item \texttt{gloppy-persimmon-ferret-3e42e8e14be0-20220716-190015} - \cavetask episode finishes in under 2 seconds with an ESC press, likely a misclick.
    \item \texttt{squeaky-ultramarine-chihuahua-bbf328311fb8-20220726-132144} - \cavetask episode where the player spawns, then falls into a ravine and dies in less than 5 seconds.
    \item \texttt{gloppy-persimmon-ferret-1d8dcc4e2446-20220716-144806} - successful \cavetask episode, where the player finds a cave and hits ESC in under 3 seconds.
    \item \texttt{pokey-cyan-spitz-62e7b7415aaf-20220714-093544} - \cavetask episode where the video lasts longer than the 3 minute time limit, but the action labels only cover 20 seconds.
    \item \texttt{whiny-ecru-cougar-f153ac423f61-20220712-192050} - \cavetask episode, where the player seems to just play Minecraft, making stone tools and such, instead of finding a cave.
    \item \texttt{shabby-pink-molly-*} - a total of 67 \cavetask episodes, where the player finds a cave, but does not press ESC to finish the episode. Instead the player proceeds to explore the cave, then exits it and goes looking for other caves.
    \item \texttt{thirsty-lavender-koala-479e09882ca6-20220717-203846} - \waterfalltask episode, one of several, where the player finishes the waterfall, and stares at it for 3 minutes to timeout instead of pressing ESC.
\end{itemize}

These were the outliers we found by sorting the data based on various metrics and checking the extremes. They are rare, on the order of tens of episodes total in a dataset with over 13,000 episodes. It would add up to a total of less than 1\%. Although we believe that these rare outliers would not influence training outcomes, we recommend that users of this dataset either remove this data from the training set or more carefully check these episodes before including them.

\subsubsection{Video Encoding Differences}

We also noticed that the codecs used to encode the videos were not always the same, likely due to subtle differences in the systems contractors used to play the game to generate the data. Codecs are how the data in the videos gets compressed and decompressed.
Roughly 88\% use \texttt{H.264 (Constrained Baseline Profile)}, while most of the remaining ones use \texttt{H.264 (High Profile)}. Also, the bitrate of the videos varies. Most bitrates are roughly 4 Mbps, but there are a few outliers on both ends of the distribution. While these differences are imperceptible to the human eye, the training algorithms might pick up on them. Having different encodings has both benefits and issues for training algorithms. The benefits are robustness, adaptability and real-world applicability. The issues include biased training data and increased computational complexity.
\section{Evaluation Dataset Details}
\label{sec:eval_appx}

This section contains details about the Evaluation Dataset. 
In this section, we hope to provide enough detail such that our evaluation pipeline can be easily reproduced. 

\subsection{Human Evaluation Details}
\label{appx:human_evaluation}

\smchange{
We estimated that one answer would take 15 minutes. 
We paid 3.75 USD per HIT for a total of 15 USD per hour.
In actuality, workers took 5.12 minutes on average to complete each evaluation, meaning the pay was closer to 43.95 USD per hour.
We also provided bonuses to MTurk workers that helped us debug issues with the form during the evaluation.
In total, we spent 14,849.16 USD on collecting this data.
}

\paragraph{MTurk Task Description}
\smchange{
The human judges viewed the following task description on MTurk.
\begin{formal}
In this questionnaire, you will watch videos of different players completing tasks in Minecraft, and your task is to judge which of the players is more successful at completing the task. This will take roughly 15 minutes of your time.
\end{formal}
}

\subsubsection{Qualification Criteria}
\smchange{We set the following criteria for selecting human judges. 
To preview the task, the judges must have had a $99\%$ or greater HIT accept rate and a minimum of $10,000$ completed HITs on MTurk. 
If the judges had these qualifications, they then took an 8-question Minecraft validity test to confirm that they had at least basic knowledge of Minecraft.
We define passing this test as reaching $65\%$ or greater.
This means that, for all checkboxes, the user correctly checked or unchecked at least $65\%$ of the boxes.
}

\paragraph{The Minecraft Qualification Test} For the purpose of completeness, we include this test.
In this section, we detail the questions that we asked the human judges.
To enable this questionnaire to be deployed, we have included deployable and more user-friendly versions of the quiz in our Github repository.
Specifically, we provide a text file and an XML file. 
The former enables the questions to be more readily copied and pasted into new formats, while the latter slots easily into the MTurk UI. 
Before answering the quiz questions, the human judges viewed the following prompt:
\begin{formal}
This is a Qualification Test to demonstrate your familiarity with Minecraft. 
This Qualification will enable you to accept HITs released by the BASALT team in relation to the NeurIPS 2022 BASALT Competition.
\end{formal}

\begin{figure}[htbp]
    \centering
    \begin{subfigure}[b]{0.45\textwidth}
        \includegraphics[clip,trim=1cm 7cm 1cm 2cm, scale=0.4, page=1]{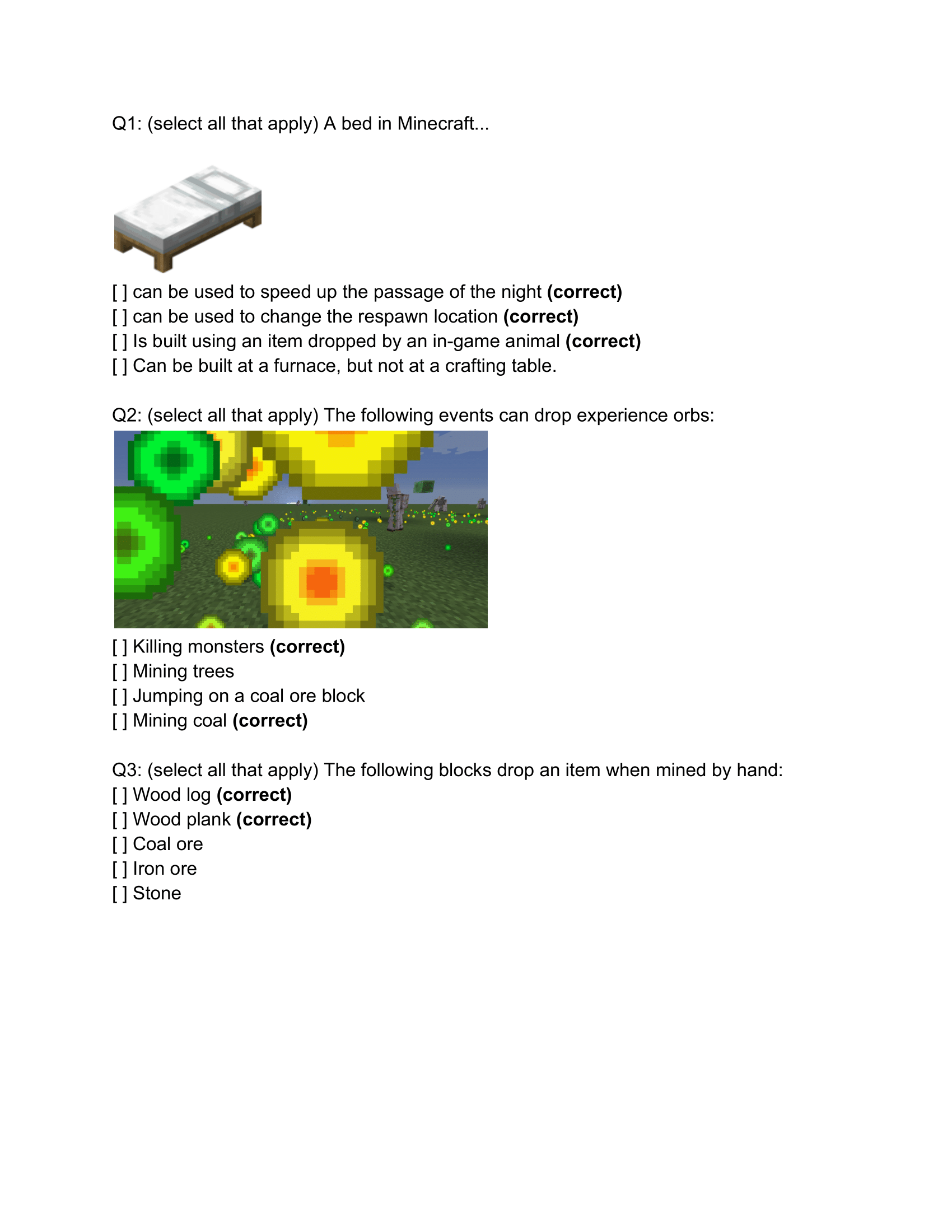}
        \caption{Page 1 of Minecraft Qualification Test}
        \label{fig:subfig1}
    \end{subfigure} 
    \hfill
    \begin{subfigure}[b]{0.45\textwidth}
        \includegraphics[clip,trim=1cm 6.9cm 1cm 2cm, scale=0.4, page=1]{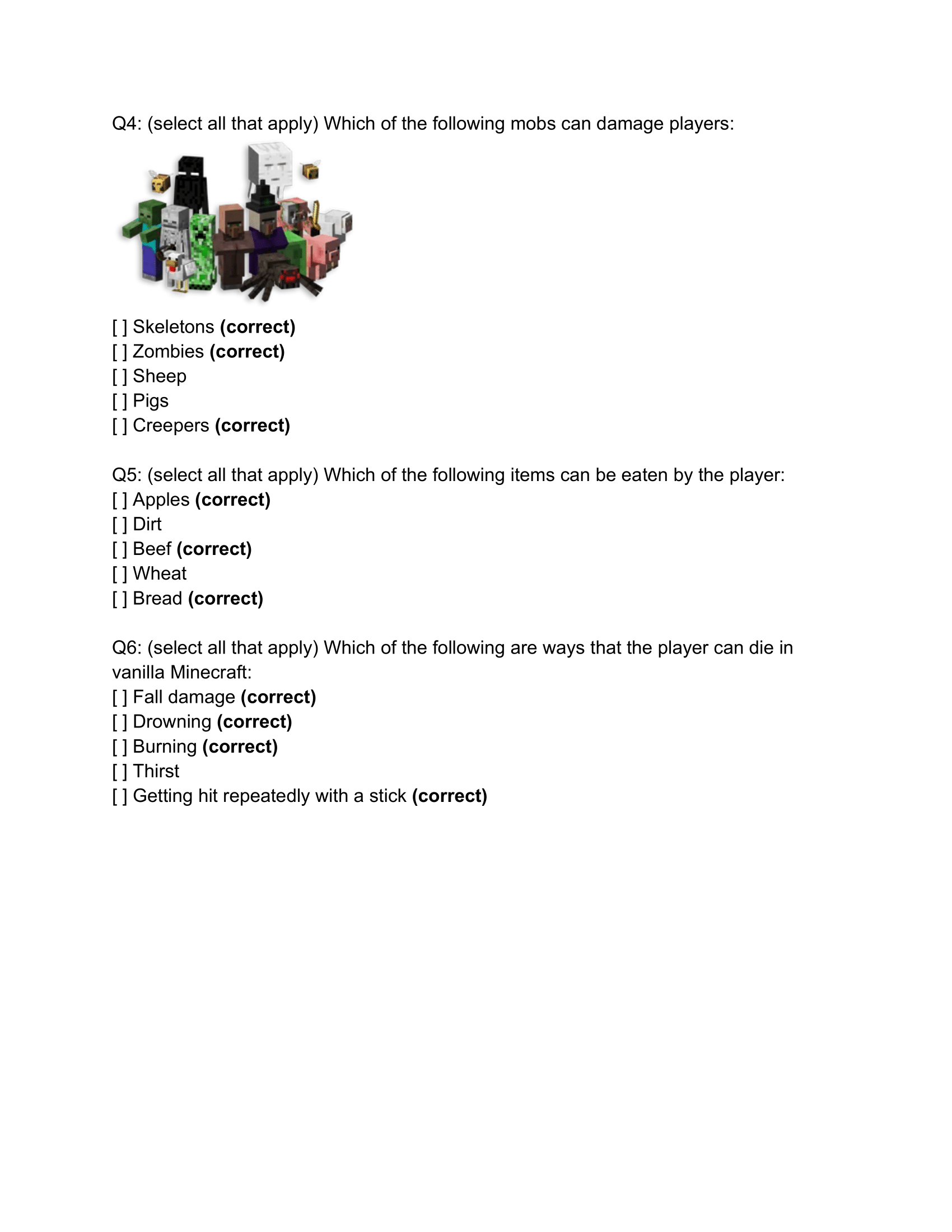}
        \caption{Page 2 of Minecraft Qualification Test}
        \label{fig:subfig2}
    \end{subfigure}
    \\ 
    \begin{subfigure}[b]{0.45\textwidth}
        \includegraphics[clip,trim=1cm 5cm 1cm 2cm, scale=0.4, page=1]{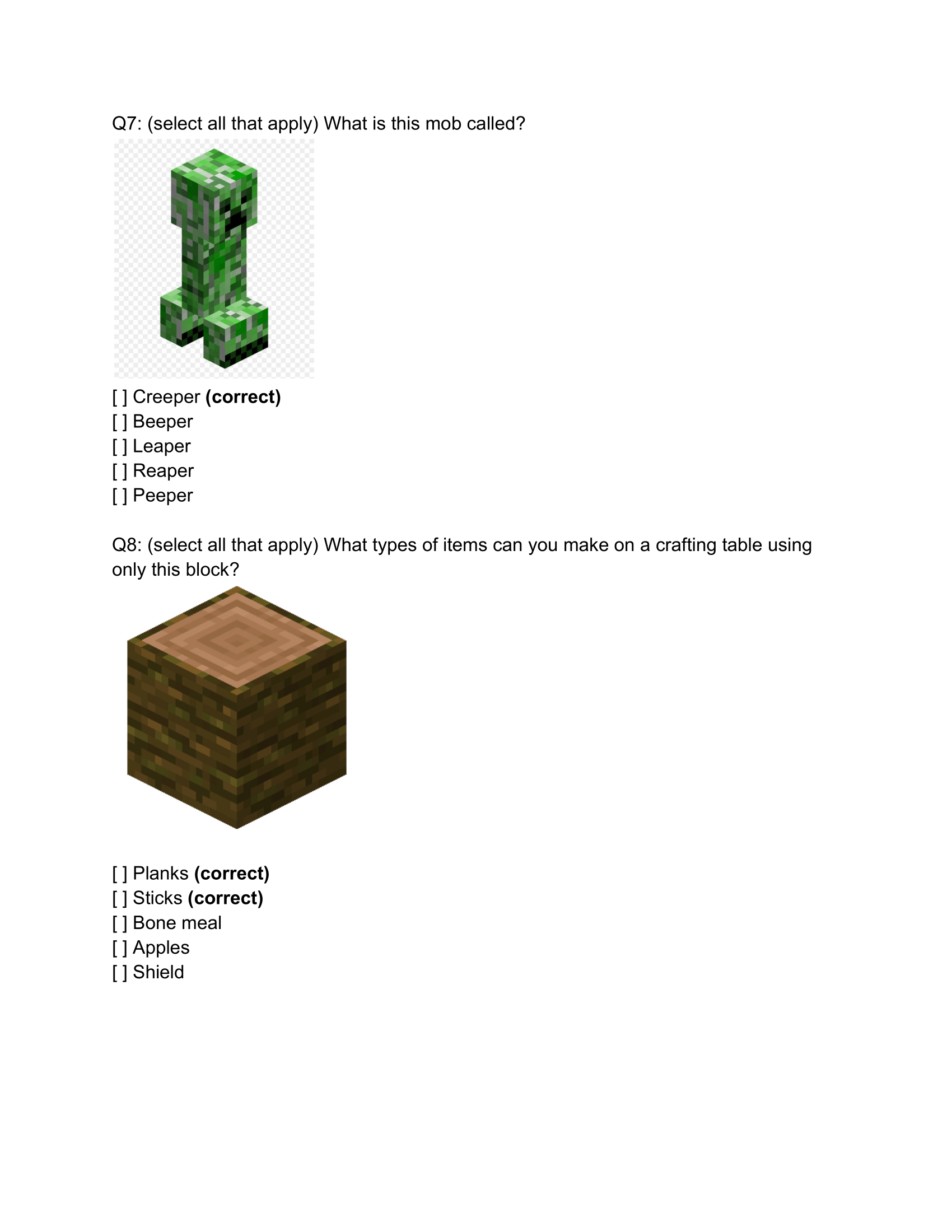}
        \caption{Page 3 of Minecraft Qualification Test}
        \label{fig:subfig3}
    \end{subfigure}
    \hfill 
    \begin{subfigure}[b]{0.45\textwidth}
        \includegraphics[clip,trim=1cm 5cm 1cm 2cm, scale=0.4, page=1]{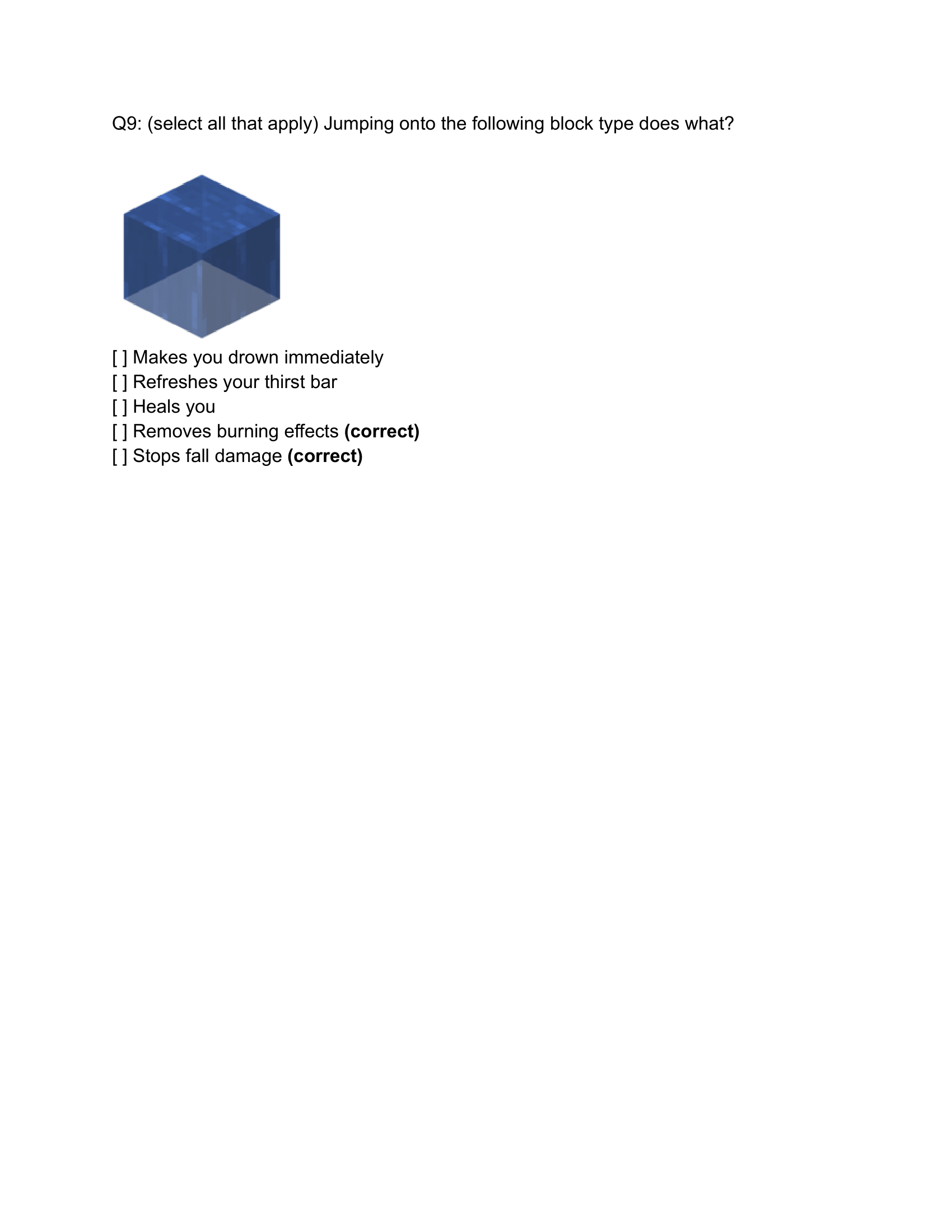}
        \caption{Page 4 of Minecraft Qualification Test}
        \label{fig:subfig4}
    \end{subfigure}
    \caption{Minecraft Qualification Test, pages 1-4 of 4 total.}
    \label{fig:mcraft_quiz}
\end{figure}

After reading this prompt, the human judges proceeded to the Minecraft Qualification Test.
\Cref{fig:mcraft_quiz} shows the questions and candidate responses in the quiz that the human judges answered to assess their knowledge of Minecraft. These questions were generated by a member of our team who has extensive experience playing Minecraft.

\subsubsection{Protocol}
\smchange{Each time a human judge accepted one of our HIT, they were directed to the AICrowd site that collects their answer.
Upon opening the webpage, the task was chosen at random from the four available ones. 
To choose the two videos to show the judge, we aimed to maximally reduce uncertainty by using the TrueSkill ranking of agents.
We ensure that the compared videos are always in the same Minecraft seed (i.e., the same starting location and surroundings).
}

\paragraph{MTurk Task Instructions}
\smchange{
Below is the full text of the MTurk task instructions given to participants.
}
\begin{formal}
Welcome to the MineRL BASALT 2022 evaluation questionnaire! This will take roughly 15 minutes of your time.
Requirements: Knowledge of Minecraft (at least of 5 hours of gameplay time with Minecraft).

In this questionnaire, you will watch videos of different players completing tasks in Minecraft, and your task is to judge which of the players is more successful at completing the task.

Videos are shown in pairs, and your task is to select which one of the two is better at solving the task. The page shows the task description. You are also given a set of more specific questions which may help you decide which of the two players completes the task better.

You may refer to the Example Videos section, for some examples of what are considered as good executions of the said task, and why.

Your answers will be used in the following ways:

To rank the solutions in the MineRL BASALT 2022 competition.
The answers will be included in the final report of the competition.
The answers may be shared publicly to support the research.
No personal information is collected.

You may complete the same questionnaire multiple times, but may be asked to judge players completing a different task. So please ensure that you take note of the Task you are submitting the responses for.
\end{formal}

\begin{figure*}[htbp]
    \centering
    \includegraphics[width=1.0\textwidth]{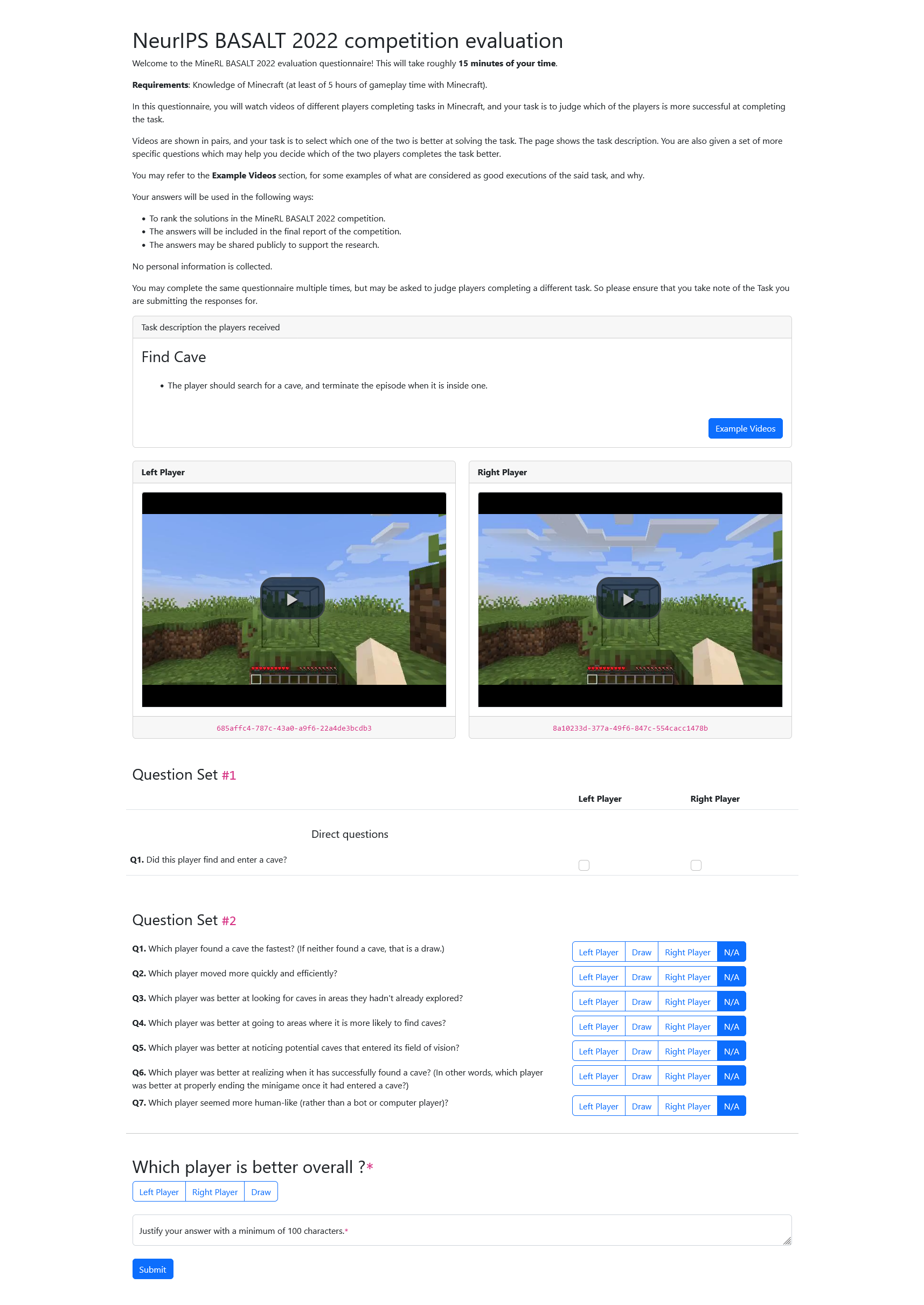}
        \caption{{Screenshot of the page that the human judges viewed when assessing the agent behavior for the \texttt{FindCave} task.} 
    }
    \label{fig:find-cave-sshot}
\end{figure*}

\begin{figure*}[htbp]
    \centering
    \includegraphics[width=1.0\textwidth]{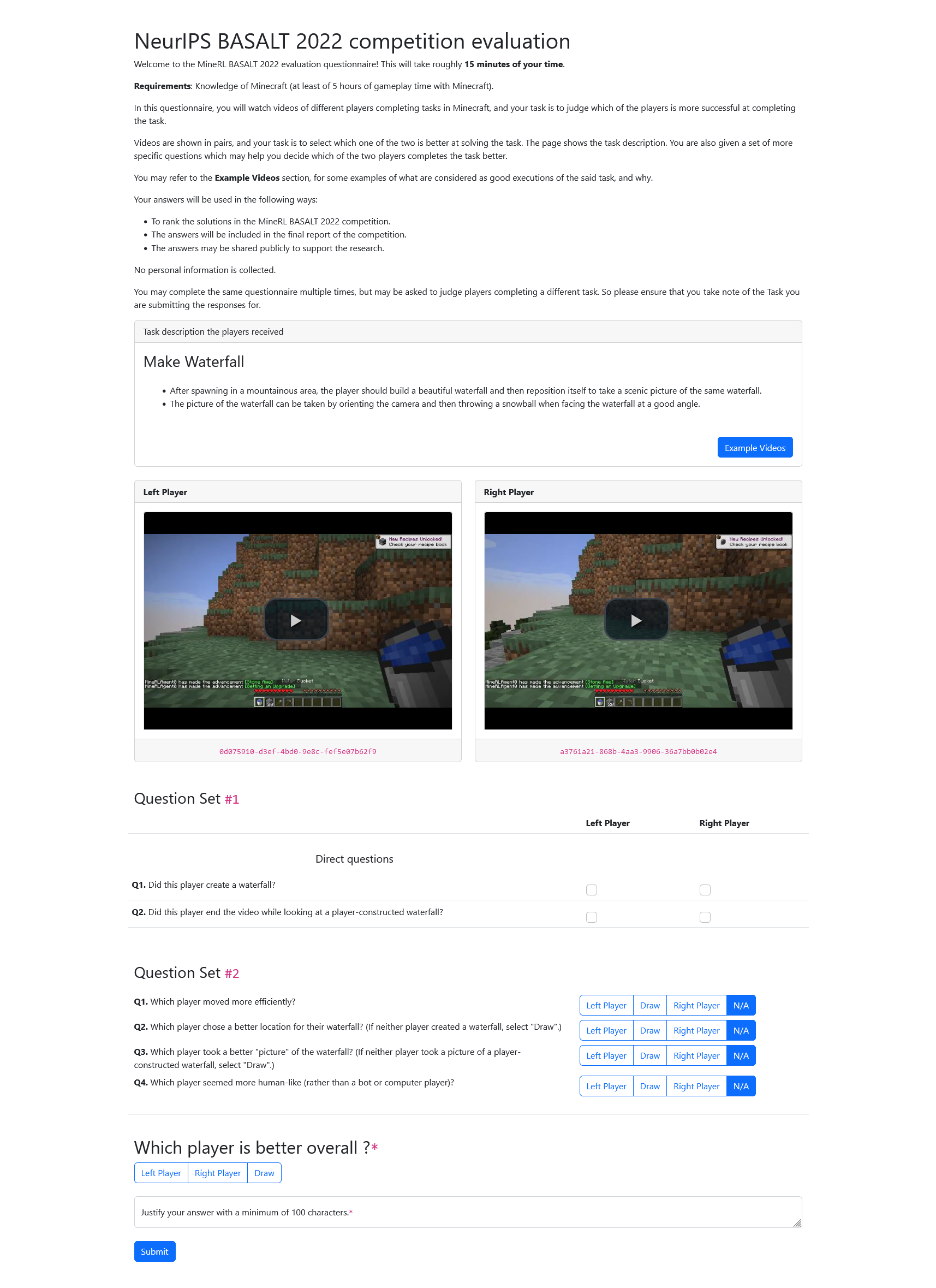}
        \caption{{Screenshot of the page that the human judges viewed when assessing the agent behavior for the \texttt{MakeWaterfall} task.} 
    }
    \label{fig:make-waterfall-sshot}
\end{figure*}

\begin{figure*}[htbp]
    \centering
    \includegraphics[width=1.0\textwidth]{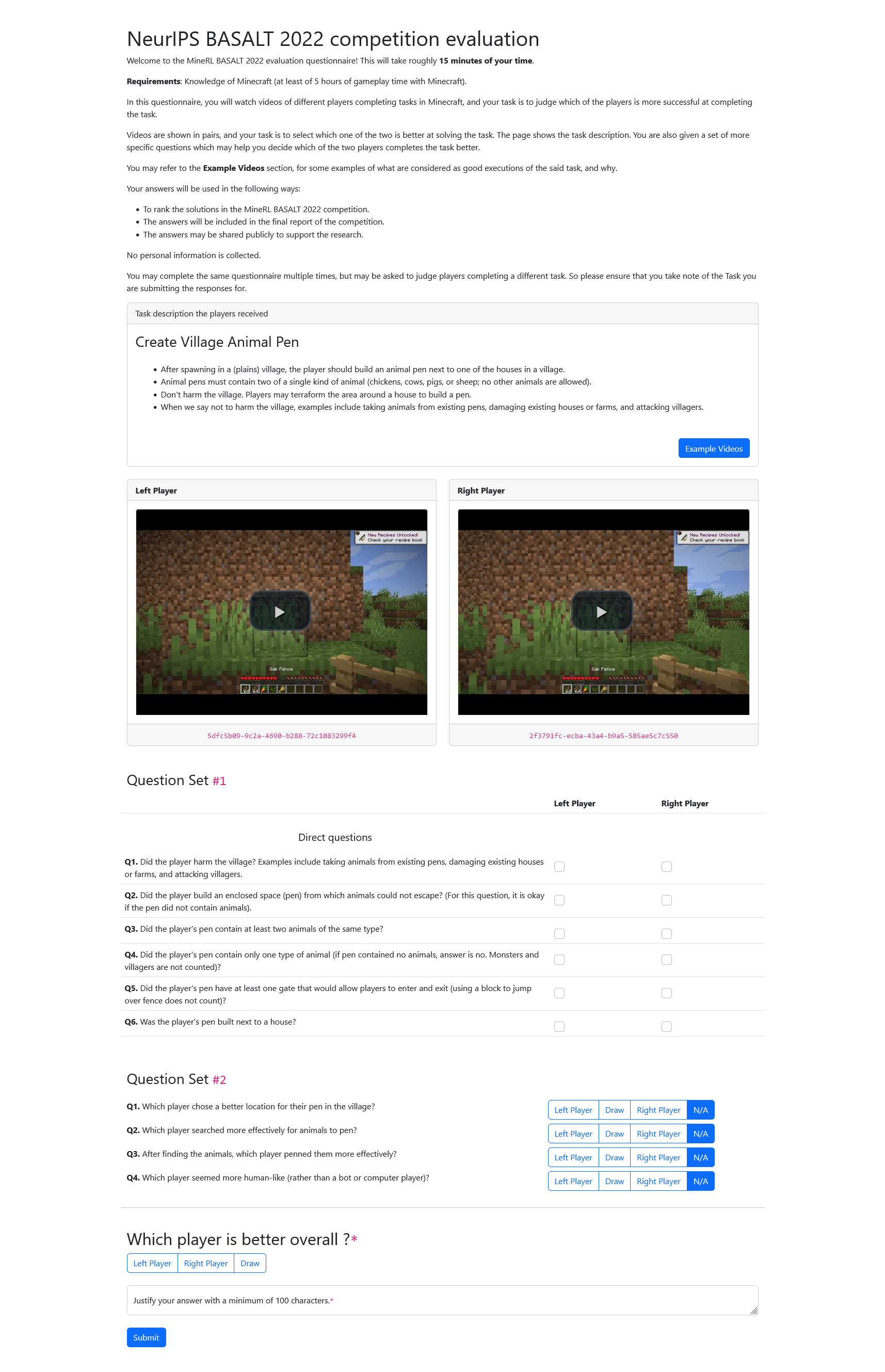}
        \caption{{Screenshot of the page that the human judges viewed when assessing the agent behavior for the \pentaskfull task.} 
    }
    \label{fig:create-pen-sshot}
\end{figure*}

\begin{figure*}[htbp]
    \centering
    \includegraphics[width=1.0\textwidth]{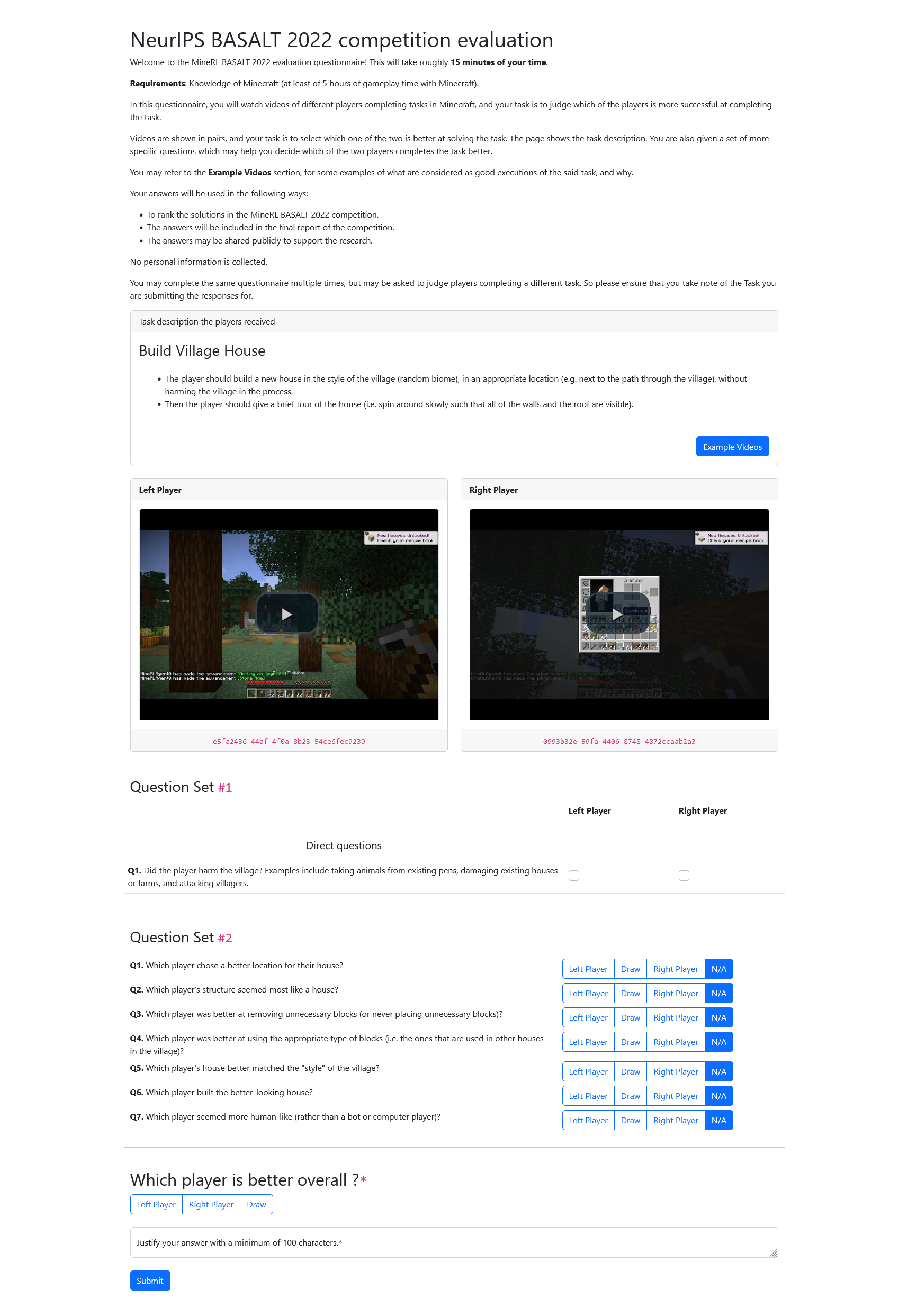}
        \caption{{Screenshot of the page that the human judges viewed when assessing the agent behavior for the \housetaskfull task.} 
    }
    \label{fig:build-house-sshot}
\end{figure*}

\paragraph{Task Details}
\smchange{
Each task contained some common questions and some different questions.
For all tasks, the human judges were asked to evaluate which player was better overall and which player appeared more human-like. 
Some tasks contained more task-specific questions than others.
For example, \housetask contained a single question about whether the players harmed the villagers; in contrast, \pentaskfull contained six questions to determine proper task completion, such as whether the pen contained at least two animals of the same type.
\Cref{fig:build-house-sshot,fig:create-pen-sshot,fig:find-cave-sshot,fig:make-waterfall-sshot} show screenshots of the form for each task that the human judges saw.
}

\section{Additional Analyses of \evaldataset}
\label{appx:eval_dataset_analyses}

This section presents the full results of the analyses performed to supplement the findings discussed in \Cref{sec:analysis_evaluation}. 
We decompose the analyses in two ways: task-based and agent-based.
\Cref{appx:task_eval_dataset_analyses} presents the task-based analyses; \Cref{appx:agent_eval_dataset_analyses} presents the agent-based analyses.

\subsection{Task-Based Decomposition}
\label{appx:task_eval_dataset_analyses}
We first present details about the analysis when we decompose responses by \textit{task}.

\begin{table}[t]
    \centering
    \begin{tabular}{lrr}
    \toprule 
    Task     &  Average Time & Total Time \\
    \midrule 
     \cavetasknospace    & $300.47 \pm 271.10$ & $216,941.18$\\
     \waterfalltask & $295.86 \pm 270.99$ & $201,778.77$ \\
     \pentaskfull & $319.52 \pm 255.50$ & $292,041.62$\\
     \housetaskfull & $375.37 \pm 315.20$ & $274,393.78$\\
     \bottomrule 
    \end{tabular}
    \caption{{More specific timing data of human evaluations in seconds, decomposed per task.}
    Error values are standard error.
    A total of around $274$ human hours were needed for evaluation. 
    }%
    \label{tab:evaluation_timing_detail}
\end{table}
\paragraph{Timing Information}
We present more detailed timing information in \Cref{tab:evaluation_timing_detail}.
This is a more detailed view of the Hours column in \Cref{tab:evaldata_stats} in the main paper.
Here, the average time means the average amount of time an evaluator took to complete a single evaluation, in seconds.
The total time is computed by adding up the number of seconds taken by all contractors to complete that type of evaluation.
On average, the human evaluators took the longest to evaluate the \housetaskfull task. 
However, there is a very high standard error, indicating that the time taken by the MTurk workers was highly variable, regardless of the task.
Overall, evaluators spent the most time on the \pentaskfullnospace. 

\begin{table}[t]
    \centering
    \begin{tabular}{lcccccc}
    \toprule 
     & \multicolumn{2}{c}{Length} &\multicolumn{3}{c}{Sentiment} \\
    Agent & Characters & Words & Positive & Neutral & Negative \\ 
    \midrule 
    \cavetasknospace & $210.46 \pm 3.82$ & $38.71 \pm 0.71$ & $79.64$\% &$6.65$\% & $13.71$\% \\
    \waterfalltask & $217.96 \pm 4.18$ & $38.76 \pm 0.79$ &$76.10$\% & $7.33$\% & $16.57$\% \\
    \pentaskfull  & $197.83 \pm 3.22$ & $35.85 \pm 0.60$ & $56.67$\% & $10.83$\% & $32.49$\% \\ 
    \housetaskfull & $205.73 \pm 4.23 $ & $36.82 \pm 0.77$ & $62.79\%$ & $9.58$\% & $27.63$\% \\
\bottomrule
    \end{tabular}
    \caption{Details about the justification provided for choosing a particular agent as the best one, decomposed per task. 
    We calculate the per-task average length of the response (characters and words) along with the standard error.
    We also report the percent of positive, neutral, and negative sentiments of these responses.
    This is a more detailed view of the Sentiment columns in \Cref{tab:eval_agents} in the main paper.}
    \label{tab:eval_justification_details_tasks}
\end{table}

\paragraph{Length of Justification}
We analyzed the length of the justifications provided by the MTurk workers for why they selected an agent as being the best at accomplishing the task.
We analyzed both the number of words and characters in the justification. 
We present the results in \Cref{tab:eval_justification_details_tasks}.
We observe that, on average, the human judges tended to dedicate the most characters to \waterfalltask and the least characters to \pentaskfullnospace.
They also tended to dedicate the most words in their responses about the \cavetask and \waterfalltask and the least number of words to \pentaskfullnospace.
We note that \pentaskfull contains more quantitative additional questions than the other tasks (4 vs. 2, the next highest). 
For example, the human judges were asked to identify which agent penned the correct amount and type of animals, among other things.
These additional questions may have enabled the judges to spend less effort explaining their choice since many of the factors were already captured by other questions. 

\begin{table}[t]
\centering
\begin{tabular}{@{}lrrr@{}}
\toprule
& & & Significance After \\
Comparison & Chi-square & p-value & Bonferroni Correction \\
\midrule
FindCave vs MakeWaterfall & 2.69 & 0.26 & No \\
\colorbox{pink}{FindCave vs CreateVillageAnimalPen} & 98.49 & 4.10e-22 & Yes \\
\colorbox{pink}{FindCave vs BuildVillageHouse} & 52.31 & 4.38e-12 & Yes \\
\colorbox{pink}{MakeWaterfall vs CreateVillageAnimalPen} & 66.37 & 3.87e-15 & Yes \\
\colorbox{pink}{MakeWaterfall vs BuildVillageHouse} & 30.50 & 2.38e-07 & Yes \\
CreateVillageAnimalPen vs BuildVillageHouse & 6.35 & 0.04 & No \\
\bottomrule
\end{tabular}
\caption{Results from Chi-square tests between tasks analyzing the sentiment of the free-form responses. Significant differences in sentiment distribution are emphasized with a \colorbox{pink}{pink highlight}.}
\label{tab:eval_justification_sentiment_bonf}
\end{table}

\paragraph{Sentiment of Justification}
We present more detailed information about the task-decomposed sentiment analysis noted in the Response Sentiment columns of \Cref{tab:evaldata_stats}. 
For completeness, we include the distribution of sentiments in \Cref{tab:eval_justification_details_tasks}.
As mentioned in the main paper, we conducted a Chi-square test of independence to examine the relationship between task and sentiment classification. 
The relation between these variables was found to be significant, $X^2(6,N=3049) = 132.21, p < .001$.
As a result, we conducted Bonferroni-corrected pairwise Chi-square tests to elucidate which of the distributions were significantly different. 
For completeness, the result of these tests is presented in \Cref{tab:eval_justification_sentiment_bonf} for completeness.
Perhaps the most interesting takeaway is that the responses for the easier tasks appear to exhibit higher positive sentiments than those for the more challenging tasks.
One explanation for this not mentioned in the main text may be that, for more challenging tasks, both agents may have exhibited some issues that were identified by the human judges.

\paragraph{Justification Examples} We present an example of justification text provided for each of the tasks. 
We choose these examples by splitting the data by task, then randomly sampling a response. 
For \cavetasknospace, an example justification was,
\begin{formal}
They explored a lot, and seemed to go to more new places than before even the water. Left player went AFK.
\end{formal}
An example justification response from \waterfalltask was,
\begin{formal}
The player on the left couldn't even manage to climb the side of a mountain, the one on the right didn't finish the tasks but at least they seemed to know how to navigate and move around.
\end{formal}
An example justification text from \pentaskfull was,
\begin{formal}
the left player completed all the requirements, while the right player did nothing and still behaved like a bot
\end{formal}
Finally, an example justification text from \housetaskfull was,
\begin{formal}
The right player was better overall, being able to build a house, but he failed to use the appropriate type of block.
\end{formal}
We believe that all of these justifications indicate that human judges were generally familiar with Minecraft and the specific requirements for task completion.

\subsection{Agent-Based Decomposition}
\label{appx:agent_eval_dataset_analyses}

\begin{table}[t]
    \centering
    \begin{tabular}{lrrrrrr}
    \toprule 
     & \multicolumn{2}{c}{Length} &\multicolumn{3}{c}{Sentiment} \\
    Agent & Characters & Words & Positive & Neutral & Negative \\ 
    \midrule 
    Random & $196.80 \pm 4.93$ & $35.41 \pm 0.92$ &$63.20$\% &$8.31$\% & $28.49$\% \\
    Human1 & $207.06 \pm 5.66$ & $37.496 \pm 1.05$ &$91.55$\% & $2.11$\% & $6.34$\% \\
    Human2   & $214.87 \pm 6.11$ & $38.86 \pm 1.12$ & $91.77$\% & $2.88$\% & $5.35$\% \\ 
    BC-Baseline &  $201.72 \pm 5.30$ & $36.25 \pm 0.97$ & $64.96\%$ & $9.00$\% & $26.03$\% \\
    GoUp & $212.86 \pm 5.55$ & $38.67 \pm 1.02$ & $73.77\%$ & $7.53\%$ & $18.70 \%$\\ 
    UniTeam & $201.77 \pm 4.85$ & $36.35 \pm 0.92$ & $65.68\%$ & $9.38\%$ & $24.94\%$\\
\bottomrule
    \end{tabular}
    \caption{Details about the justification provided for choosing a particular agent as the best one. 
    We calculate the per-agent average length of the response (characters and words) along with the standard error.
    We also report the percent of positive, neutral, and negative sentiments of the responses.
    This is a more detailed view of the Words in Response and Response Sentiment columns in \Cref{tab:evaldata_stats} in the main paper.}
    \label{tab:eval_justification_details_agents}
\end{table}

We now present details about the analysis when we decompose responses by \textit{agent}.
Although we release the \textit{full} dataset, which includes the values for all agents, we present here only the specific values for Human1, Human2, BC-Baseline, UniTeam\footnote{\url{https://github.com/fmalato/basalt_2022_submission}}, GoUp\footnote{\url{https://github.com/gomiss/neurips-2022-minerl-basalt-competition}}, and Random.
We compute these values using only the comparisons with the other presented algorithms, not the full dataset. 

\paragraph{Length of Justification}
We analyzed the length of the justifications provided by MTurk workers. 
These results are presented in the Length column of \Cref{tab:eval_justification_details_agents}.
On average, the MTurk workers used both the most words and characters (38.86 words, 214.87 characters) when describing their rationale when Human2 was involved in the pair being compared.
In contrast, when the Random agent was involved, the MTurk workers used the least words and characters (35.41 words and 196.80 characters) when explaining their rationale.
We believe that this may be due to the relative competency of the agents: because it is easier to identify the Random agent as less skilled and the humans as more skilled, the human judges may require less justification for their selection.
However, we note that the difference between these agents for both words and characters (3.45 words and 18.07 characters) is relatively small overall. 

\paragraph{Sentiment of Justification}
\begin{table}[t]
\centering
\begin{tabular}{lrrr}%
\toprule
 & & & Significance After  \\
Comparison & Chi-square & p-value & Bonferroni Correction \\
\midrule
GoUp vs UniTeam & 6.37 & 0.041 &  No \\
GoUp vs BC-Baseline & 7.50 & 0.024 & No \\
GoUp vs Random & 10.44 & 0.005 & No \\
\colorbox{pink}{GoUp vs Human1} & 34.10 & $3.93 \times 10^{-8}$ & Yes \\
\colorbox{pink}{GoUp vs Human2} & 31.22 & $1.66 \times 10^{-7}$ & Yes \\
UniTeam vs BC-Baseline & 0.15 & 0.928 & No\\
UniTeam vs Random & 1.33 & 0.515 & No\\
\colorbox{pink}{UniTeam vs Human1} & 62.97 & $2.12 \times 10^{-14}$ & Yes \\
\colorbox{pink}{UniTeam vs Human2} & 56.94 & $4.31 \times 10^{-13}$ & Yes \\
BC-Baseline vs Random & 0.60 & 0.740 & No\\
\colorbox{pink}{BC-Baseline vs Human1} & 64.77 & $8.64 \times 10^{-15}$ & Yes \\
\colorbox{pink}{BC-Baseline vs Human2} & 58.76 & $1.74 \times 10^{-13}$ & Yes \\
\colorbox{pink}{Random vs Human1} & 68.25 & $1.51 \times 10^{-15}$ & Yes \\
\colorbox{pink}{Random vs Human2} & 62.44 & $2.77 \times 10^{-14}$ & Yes \\
Human1 vs Human2 & 0.53 & 0.767 & No \\
\bottomrule
\end{tabular}
\caption{Results from Chi-square tests between teams analyzing the sentiment of the free-form responses. Significant differences in sentiment distribution are emphasized with a \colorbox{pink}{pink highlight}.}
\label{tab:agent_sentiment_chi_square_results}
\end{table}

We now present the full statistics for our agent-based sentiment analysis.
The sentiments are captured in the Sentiment columns of \Cref{tab:eval_justification_details_agents}. 
As mentioned in the main paper, we conducted Bonferroni-corrected pairwise Chi-square tests to elucidate which of the agent types exhibited different distributions of sentiment.
\Cref{tab:agent_sentiment_chi_square_results} shows the result of this analysis.
We find that any comparisons that include either of the human agents, except for when they are pitted against one another, exhibit significant differences in sentiment distribution.
The responses had the highest positive sentiment for the two human agents and the least positive sentiment for the Random agent.
However, the difference in sentiment distribution of sentiment was statistically insignificant when comparing Random with BC-Baseline or UniTeam, meaning that the sentiment was similarly negative.

\paragraph{Factors}
\begin{sidewaystable}[t]
    \centering
    \begin{tabularx}{\textwidth}{p{3cm}Xp{5.3cm}}
    \toprule 
     Task & Question & Factor \\
     \midrule 
     \cavetask & \textbf{Quantitative} & \\
     & Did this player find and enter a cave? & Found Cave \\ 
     & \textbf{Qualitative} \\
     & Which player found a cave the fastest? (If neither found a cave, that is a draw) & Found Cave Faster \\
     & Which player moved more quickly and efficiently? & More Quick and Efficient Movement \\ 
     & Which player was better at looking for caves in areas they hadn't already explored? & Better Cave Search in Unknown Areas \\
      & Which player was better at going to areas where it is more likely to find caves? & Better Navigation to Cave Areas \\ 
      & Which player was better at noticing potential caves that entered its field of vision? & Better Cave Detection \\
      & Which player was better at realizing when it had successfully found a cave? (In other words, which player was better at properly ending the minigame once it had entered a cave?) & Better Cave Perception \\
      & Which player seemed more human-like (rather than a bot or computer player)? & More Human Like \\
      \hline 
      \waterfalltask & \textbf{Quantitative} & \\ 
      & Did this player create a waterfall? & Created Waterfall \\
      &Did this player end the video while looking at a player-contructed waterfall? & Looked at Waterfall \\
      & \textbf{Qualitative} & \\
      & Which player moved more efficiently? & More Quick and Efficient Movement \\ 
      & Which player chose a better location for their waterfall? (If neither player created a waterfall, select ``Draw''.) & Better Location for Waterfall \\
      & Which player took a better ``picture'' of the waterfall? (If neither player took a picture of a player-constructed waterfall, select ``Draw'') & Better Picture of Waterfall \\
      & Which player seemed more human-like (rather than a bot or computer player)? & More Human Like \\
      \hline 
      \housetaskfull & \textbf{Quantitative} &  \\
      &Did the player harm the village? Examples include taking animals from existing pens, damaging existing houses or farms, and attacking villagers. & Least Village Harm \\
      & \textbf{Qualitative} \\
      &Which player chose a better location for their house? & Better House Location \\
      &Which player's structure seemed most like a house?& More House-Like Structure \\
      &Which player was better at removing unnecessary blocks (or never placing unnecessary blocks)?& Intentional Block Placing \\ 
      &Which player was better at using the appropriate type of blocks (i.e., the ones that are used in other houses in the village)? & Appropriate Block Placing \\ 
      &Which player's house better matched the ``style'' of the village? & Better Style Matching \\ 
      &Which player built the better-looking house? & More Attractive House \\
      & Which player seemed more human-like (rather than a bot or computer player)? & More Human Like \\
      \hline 
      \pentask &\textbf{Quantitative} & \\ 
      &Did the player harm the village? Examples include taking animals from existing pens, damaging existing houses or farms, and attacking villagers. & Least Village Harm \\
      &Did the player build an enclosed space (pen) from which animals could not escape? (For this question, it is okay if the pen did not contain animals). & Better Enclosed Space \\
      &Did the player's pen contain at least two animals of the same type? & Correct Number of Animals \\
      &Did the player's pen contain only one type of animal (if pen contained no animals, answer is no. Monsters and villagers are not counted)? & Correct Type of Animals \\
      &Did the player's pen have at least one gate that would allow players to enter and exit (using a block to jump over fence does not count)? & Proper Exit \\
      &Was the player's pen built next to a house? & Next to House \\
      & \textbf{Qualitative} \\
      &Which player chose a better location for their pen in the village?& Better Location \\
      &Which player searched more effectively for animals to pen?& More Effective Search \\
      &After finding the animals, which player penned them more effectively?& More Effective Penning \\
      & Which player seemed more human-like (rather than a bot or computer player)? & More Human Like \\
      \bottomrule 
    \end{tabularx}
    \caption{Mapping for each task from question asked of human evaluators to presented attribute.}
    \label{tab:attribute_mapping}
\end{sidewaystable}

Here we present the mapping of per-task questions to factors, which were presented in \Cref{fig:human_eval_questions} from the main text.
We map each question to an attribute. 
\Cref{tab:attribute_mapping} shows this mapping.

\section{Discussion}
\label{appx:discussion}
Here we provide a longer discussion about the limitations of our work and suggest future work that stems from these limitations (\Cref{sec:limitations}).
We conclude by discussing the potential societal impacts of this work (\Cref{sec:societal_impact}). 

\subsection{Limitations and Future Work}
\label{sec:limitations}

\paragraph{Low-Level Data Discrepencies}
As with any dataset, ours is not without issues. 
We discussed some limitations of the Demonstrations Dataset in a previous section of the appendix. 
We found that it is often challenging to delineate episode boundaries with data recording tools and splitting methods.
In contrast, with games like Atari, episode boundaries are easily provided by the simulator.
Furthermore, since the data is temporally-extended and human-generated, there are idiosyncrasies that will necessarily arise. 
We emphasize that rather than deploying the dataset and making users find these issues, we dedicated time to investigating these issues and proposing solutions. 

\paragraph{Demographic Information}
In both studies, we did not collect any additional demographic information about participants. 
On one hand, this lack of personally-identifiable information decreases the likelihood of deanonymization of the participants, which is important even in a study with relatively low stakes such as this one. 
On the other hand, there is a missed opportunity to critically analyze the influence of demographic or other factors on either the produced demonstrations or the assessments. 
Future work might include some of these demographic details to produce a more in-depth analysis of the assessments. 

\paragraph{Fine-Grained Assessment Details}
In the assessment of the agents, there are still even more fine-grained details that could be assessed. 
For example, how do we define human likeness? 
Is the goal to produce agents that behave as though they are controlled by a person or like they are embodied in the real world? 
What \textit{type} of human are they similar to: a novice Minecraft player? 
An expert? 
For many of these comparative questions, we can perform this finer-grained assessment, depending on the factors that are most relevant to the study. 
Continuing with the human likeness example, future work investigating human likeness may want to substitute some of the non-human-like comparative questions with more detailed questions about human likeness to gain a deeper understanding of assessments of human likeness in this setting. 
This analysis could also be combined with recent work on the Human Navigation Turing Test (HNTT) and Automated Navigation Turing Test (ANTT)~\citep{devlin2021navigation} but applied to a much more complex setting: completing fuzzy tasks in Minecraft.

\paragraph{English-Language Focus}
One aspect of our study that offers both a specific focus and an avenue for exploration is the use of English-language descriptions for tasks. 
Incorporating multiple languages into the dataset would make the benchmark more globally applicable and uncover new insights about how language and cultural context influence human feedback.

\subsection{Societal Impact}
\label{sec:societal_impact}
We believe that publicly sharing these datasets helps foster a more open and inclusive AI research community in which resources are more broadly accessible.
However, there are some potential negative societal impacts that we have considered. 
There is a small (but highly unlikely) chance that we missed some inappropriate or harmful symbols or language contained in the dataset. 
Although the competition-driven nature of the tasks may foster faster progress toward agents that can better learn from human feedback, releasing all data and code may lead to researchers overfitting on the subset of tasks. 
This may result in a false signal of progress. 
There is therefore a need to utilize or develop more non-programmatic tasks in Minecraft~\citep{fan2022minedojo} and other domains.

\end{document}